%% file: main.tex
\documentclass[lettersize,journal]{IEEEtran}
\usepackage{amsmath,amsfonts,amssymb}
\usepackage{algorithm}
\usepackage{algpseudocode}
\usepackage{array}
\usepackage{booktabs}
\usepackage{multirow}
\usepackage{graphicx}
\usepackage{tikz}
\usetikzlibrary{arrows.meta, positioning, calc}
\usepackage{textcomp}
\usepackage{url}
\usepackage{xcolor}
\usepackage{cite}
\usepackage[hidelinks]{hyperref}
\usepackage{dblfloatfix}

\usepackage{eso-pic}
\begin{document}
\AddToShipoutPictureBG*{%
  \AtPageLowerLeft{%
    \put(\LenToUnit{\dimexpr(\paperwidth-\textwidth)/2\relax},\LenToUnit{0.09in}){%
      \parbox[b]{\textwidth}{\fontsize{6.5}{7.5}\selectfont\raggedright
        \copyright~2026 IEEE. Personal use of this material is permitted. Permission from IEEE must be
        obtained for all other uses, in any current or future media, including reprinting/republishing
        this material for advertising or promotional purposes, creating new collective works, for resale
        or redistribution to servers or lists, or reuse of any copyrighted component of this work in
        other works.
        This is the accepted version of: Y.~Jiang, G.~Deng, M.~Xu, C.~Ruan, J.~Zhao, and L.~Huang,
        ``GeoSelect: Spatial-Program Execution for Training-Free Referring Remote Sensing Image
        Segmentation,'' \emph{IEEE Trans. Geosci. Remote Sens.}, 2026, doi: 10.1109/TGRS.2026.3734378.
        The published version is available at {\urlstyle{same}\url{https://ieeexplore.ieee.org/document/11693937}}.}}}}

\title{GeoSelect: Spatial-Program Execution for\\
Training-Free Referring Remote Sensing Image Segmentation}

\author{Yuhang~Jiang, Guohui~Deng, Miaozhong~Xu, Chao~Ruan, Jinling~Zhao,~\IEEEmembership{Member,~IEEE,}
and Linsheng~Huang%
\thanks{Yuhang Jiang, Guohui Deng, Chao Ruan, Jinling Zhao, and Linsheng Huang are with the School of
Internet, Anhui University, Hefei, China. Guohui Deng is also with the National Engineering
Research Center for Agro-Ecological Big Data Analysis \& Application, Anhui University, Hefei,
China.}%
\thanks{Miaozhong Xu is with the State Key Laboratory of Information Engineering in Surveying,
Mapping and Remote Sensing, Wuhan University, Wuhan, China.}%
\thanks{Corresponding author: Guohui Deng (e-mail: dlyj0207@whu.edu.cn).}}

\markboth{IEEE Transactions on Geoscience and Remote Sensing, 2026}%
{Jiang \MakeLowercase{\textit{et al.}}: GeoSelect: Spatial-Program Execution for Training-Free RRSIS}

\maketitle

\input{sections/abstract}
\input{sections/introduction}
\input{sections/related_work}
\input{sections/problem_formulation}
\input{sections/method}
\input{sections/experiments}
\input{sections/discussion}
\input{sections/conclusion}

\appendices
\section{Synthesis Schema and Reproducibility}
\label{sec:appendix:schema}
\input{tables/tab_mapping}
The synthesiser uses a single fixed few-shot prompt, released verbatim with the code. It declares
the operator grammar of Fig.~\ref{fig:grammar}, the image-coordinate polarity (X grows
left-to-right and Y top-to-bottom, so \emph{topmost} maps to \textsc{Argmin} on $Y$ and
\emph{bottommost} to \textsc{Argmax} on $Y$), and a fixed set of language-to-program rules:
image-frame phrases (``on the right'', ``in the lower left'') map to a \textsc{Filter} anchored to
the image; ``middle''/``centre'' maps to the \textsc{Center} predicate and never to an extremum;
corner phrases map to a single diagonal predicate (``lower left'' $\to$ \textsc{LL}); positional
superlatives map to \textsc{Argmax}/\textsc{Argmin} on the axis the superlative word selects; ``$n$-th from''
maps to \textsc{Nth}; an object-relative direction (``left of the harbor'') maps to a monotone
\textsc{Filter} with a nested anchor, while \textsc{Relate} is reserved for
\textsc{Near}/\textsc{Between}/\textsc{Adjacent}/\textsc{Contains}/\textsc{Inside}; and colour,
shape, and non-comparative size adjectives stay inside the \textsc{Select} noun phrase rather than becoming
predicates. The prompt is verified disjoint from the validation and test expressions at the
string level (exact match after lowercasing and whitespace normalisation). It was developed on
RRSIS-D val and applied unchanged to RISBench. The prompt is deliberately over-specified for
safety: keeping only the prose rules (no examples), only the examples (no rules), or only four of
the fourteen examples each leaves \emph{every} prediction byte-identical on RRSIS-D val, so
the reported accuracy does not depend on prompt micro-tuning. The prompt is scaffolding, not a
contribution; the substantive ablation of the executed representation is the
fields-only-versus-full comparison (Section~\ref{sec:exp:ops}, Table~\ref{tab:risbench_ops}), not
the prompt wording. Table~\ref{tab:mapping} lists the explicit phrase-to-operator mapping.

\noindent\textbf{Grammar (EBNF).} The DSL is the typed grammar below. \textsc{Select} maps a
detector-query noun phrase to the initial scored candidate set~$\mathcal{C}$; the unary operators
map $\mathcal{C}\!\rightarrow\!\mathcal{C}$ and \textsc{And}/\textsc{Or} map
$\mathcal{C}\!\times\!\mathcal{C}\!\rightarrow\!\mathcal{C}$, so the
root returns a scored candidate set whose top-scoring box is passed to the segmenter.
\begin{center}\scriptsize
\begin{tabular}{@{}l@{\ }l@{}}
\texttt{program} & \texttt{::=\ expr} \\
\texttt{expr}    & \texttt{::=\ Select(phrase)} \\
                 & \texttt{\ |\ Filter(expr, pred, anchor)} \\
                 & \texttt{\ |\ Argmax(expr, axis)\ |\ Argmin(expr, axis)} \\
                 & \texttt{\ |\ Nth(expr, index, axis)} \\
                 & \texttt{\ |\ RestrictCount(expr, count)} \\
                 & \texttt{\ |\ Relate(expr, rel, anchor\ [, anchor])} \\
                 & \texttt{\ |\ And(expr, expr)\ |\ Or(expr, expr)\ |\ Not(expr)} \\
\texttt{anchor}  & \texttt{::=\ IMAGE\ |\ expr} \\
\texttt{pred}    & \texttt{::=\ LEFT\ |\ RIGHT\ |\ TOP\ |\ BOTTOM} \\
                 & \texttt{\ |\ UL\ |\ UR\ |\ LL\ |\ LR\ |\ CENTER\ |\ NEAR} \\
\texttt{rel}     & \texttt{::=\ NEAR\ |\ BETWEEN\ |\ ADJACENT\ |\ CONTAINS\ |\ INSIDE} \\
\texttt{axis}    & \texttt{::=\ X\ |\ Y\ |\ DIAG\_UL\_LR\ |\ DIAG\_UR\_LL} \\
\texttt{index}   & \texttt{::=\ non-negative integer (zero-based)} \\
\texttt{count}   & \texttt{::=\ positive integer} \\
\texttt{phrase}  & \texttt{::=\ detector-query noun phrase} \\
\end{tabular}
\end{center}
\noindent Here \texttt{phrase} retains colour, non-comparative size, and shape adjectives and is passed verbatim to
the detector; \texttt{Nth} takes a zero-based index and \texttt{RestrictCount} a positive count, and
\texttt{Relate} takes one anchor, or two for \textsc{Between}. Fig.~\ref{fig:grammar} summarises the
same grammar graphically.

\noindent\textbf{Worked execution examples.} Two examples trace a program from expression to the
selected box, where \textsc{Detect} returns the candidate set for each \textsc{Select} phrase.
(i)~\emph{``the rightmost tennis court''} synthesises to
$\textsc{Argmax}(\textsc{Select}(\text{``tennis court''}),\,X)$. \textsc{Select} returns the
detected tennis courts as a scored set; \textsc{Argmax} on the $x$ axis keeps the single right-most
box and zeroes the others, reducing the set to one element; that box is segmented.
(ii)~\emph{``the ship on the left near the top''} synthesises to
\begin{equation*}
\begin{aligned}
\textsc{And}\big(&\textsc{Filter}(\textsc{Select}(\text{``ship''}),\textsc{Left},\textsc{Image}),\\
&\textsc{Filter}(\textsc{Select}(\text{``ship''}),\textsc{Top},\textsc{Image})\big).
\end{aligned}
\end{equation*}
Both operands
query the same ``ship'' detections, so their candidates carry identical box identities; each
\textsc{Filter} multiplies the shared scores by its directional field, \textsc{And} takes the
per-candidate minimum, and the ship that is both on the left and near the top is selected. In each
case the output is a single-instance mask.

\noindent\textbf{Reproducibility.} The code, the frozen configuration, the complete prompt, the
per-sample IoU dumps, and the bootstrap-CI script used for all confidence intervals are publicly
available at \url{https://github.com/Avalon-S/GeoSelect}, so the reported numbers can be recomputed
without re-running detection or segmentation.

\section*{Acknowledgments}
This research was supported in part by the Open Research Fund of the National Engineering Research
Center for Agro-Ecological Big Data Analysis \& Application, Anhui University (No.~AE202501); the
Natural Science Foundation of Anhui Province (No.~2508085QD124); the Anhui Provincial Natural
Science Foundation (No.~2508085MD080); and the Anhui International Joint Research Center for Ancient
Architecture Intellisencing and Multi-Dimensional Modeling (No.~GJZZX2022KF04).

\bibliographystyle{IEEEtran}
\bibliography{GeoSelect}

\end{document}

%% file: sections/abstract.tex
\begin{abstract}
Referring remote sensing image segmentation segments the object
named by a natural-language expression in an aerial image. Existing training-free methods
resolve the expression through implicit vision--language activations or region--text
similarity, which gives weak control over the spatial, superlative, and ordinal relations
that dominate aerial referring, such as \emph{the rightmost ship} or \emph{the second court from
the left}. We propose GeoSelect, a training-free
pipeline that reframes referring as the execution of a typed spatial program. A frozen,
text-only language model synthesises the expression into a small domain-specific language, a
well-formedness checker accepts the program, and a deterministic executor runs it. The central
abstraction is a single \emph{scored candidate set} type under which every operator composes:
continuous geometric fields realise position and proximity, while
discrete set and order operators add the extremum, ordinal, top-$k$, and relational
constructions that fields alone cannot express. Execution is explicit, so every intermediate
is inspectable, and a reliability ladder degrades any failing program
to the field-only special case. GeoSelect achieves
58.86~mIoU on RRSIS-D test and 55.27~mIoU on RISBench test, more than twice the best prior
training-free method on RRSIS-D, with no referring supervision and on a single GPU. Under a fixed
detector and segmenter, explicit execution improves over implicit selectors under the same backbone;
the best-box-IoU and outcome-partition diagnostics motivate complementary tests of proposal recall and
program-path behaviour, with the program path the clearer priority on RISBench, and an exposure audit
shows comparable accuracy on the audited unseen subset. Code and configurations are available at \url{https://github.com/Avalon-S/GeoSelect}.
\end{abstract}

\begin{IEEEkeywords}
Referring image segmentation, remote sensing, training-free, neuro-symbolic reasoning,
program synthesis, vision--language models, spatial reasoning.
\end{IEEEkeywords}

%% file: sections/introduction.tex
\section{Introduction}
\label{sec:intro}
\IEEEPARstart{R}{eferring} remote sensing image segmentation (RRSIS)~\cite{rmsin}
requires instance-level disambiguation in crowded aerial scenes. Expressions frequently
specify the referent through spatial, superlative, and ordinal cues such as
\emph{the upper-right ship}, \emph{the rightmost tank}, or \emph{the second court from the
left}, where several same-class objects would otherwise be valid candidates. Resolving the
referent therefore demands that the relational content of the expression be evaluated, not
merely that the object category be recognised.

Recent supervised networks~\cite{rmsin,crobim,fianet} reach strong accuracy on RRSIS-D, but
require dense referring masks and a separate training run for each benchmark. Training-free
methods instead combine frozen pretrained components, yet the best published result on
RRSIS-D test remains far below the supervised specialists. The gap suggests that pretrained
perception alone does not solve spatial referring.

\input{figures/fig_teaser}

We study the representation of the spatial constraint (Fig.~\ref{fig:teaser}). Existing
training-free pipelines derive the referent from vision--language activations, attention, or
region--text similarity. These signals may identify the relevant object category, but do not
directly execute relations such as \emph{left of}, \emph{near}, \emph{rightmost}, or
\emph{second from the left}. In a controlled selector comparison with detection and
segmentation held fixed, explicit geometric execution outperforms
the stronger of two implicit selectors, and the gap is concentrated on the spatial and compositional expressions
and nearly disappears on the pure-attribute subset. This advantage widens sharply with same-class
clutter; overhead imagery, which packs many near-identical instances
into a single scene, makes this clutter regime the rule rather than the exception. Inter-instance
geometric selection is thus where aerial referring is hardest, and where explicit execution helps
most, while recognition-only signals are weakest.

Our starting point is a field-only special case that represents each expression as a fixed relation over
a closed lexicon of positional and proximity predicates, composes their pixel-level
fields, and ranks candidate boxes by the composed prior. This covers image-relative,
object-anchored, and attribute expressions, but it cannot express an extremum, a rank, or a
counted sub-collection: these operate on the candidate set rather than on fields over the image
plane, and they dominate the longer expressions of RISBench. GeoSelect removes this ceiling by reframing referring as explicit
spatial-program execution. A frozen, text-only language model synthesises a typed program in
a small domain-specific language; a well-formedness checker accepts or rejects it; and a deterministic
executor runs it over a single scored-candidate-set type that unifies the continuous field
algebra with discrete set and order operators. Position and proximity remain dense geometric
fields, while superlative, ordinal, and compositional expressions are synthesised from
extremum, ordinal, top-$k$, and relation operators that the closed lexicon lacked. Every
intermediate program, field, and ranking is inspectable, and an explicit reliability ladder
degrades any ill-formed or failing program to the field-only selector, so a malformed program
forfeits no answer. Although executing a program over perception modules has precedent in
natural-image referring~\cite{visprog,vipergpt}, GeoSelect is built for the aerial regime rather
than ported from it. Overhead scenes pack numerous same-class instances and complex geospatial
relations that make instance disambiguation, not recognition, the core
difficulty~\cite{rmsin,crobim}, so the bottleneck is inter-instance geometric selection; and
open-vocabulary detectors pretrained on natural scenes transfer poorly to overhead
imagery~\cite{laedino}, so candidates must come from an aerial-pretrained detector.

\noindent\textbf{Contributions.}
\begin{itemize}
\item \textbf{(C1) Referring as explicit spatial-program execution.}
  We recast training-free RRSIS as executing a typed spatial program over a single
  scored-candidate-set type. With no referring supervision and on a single GPU, the pipeline
  reaches $58.86$~mIoU on RRSIS-D test, more than twice the best prior training-free result and
  about $93\%$ of the supervised specialist RMSIN~\cite{rmsin}.
\item \textbf{(C2) Discrete operators and controlled attribution.}
  Extremum, ordinal, top-$k$, and binary-relation operators make the superlative, ordinal,
  and compositional constructions that a closed positional lexicon cannot represent expressible in
  the first place. With the detector and segmenter fixed, explicit execution outperforms the implicit
  selectors, and the best-box-IoU proxy and the outcome partition suggest proposal recall and the
  program path as complementary directions for further gains.
\item \textbf{(C3) Cross-domain accuracy and a leakage audit.}
  The single frozen configuration transfers across domains to RISBench at $55.27$~mIoU, more than
  $1.7$ times the best prior training-free method. An image-level pretraining-exposure audit
  further shows that the exposure-audited unseen subset reaches $58.4$~mIoU ($54.5$ under a
  conservative seen-set), still well above prior training-free methods; this exposure is quantified,
  not left unexamined.
\end{itemize}

Design choices are made on RRSIS-D val; the resulting configuration is frozen before any test
evaluation and applied to RISBench with no test-set recalibration.
Sections~\ref{sec:problem}--\ref{sec:exp} present the formulation, method, and experiments.

%% file: figures/fig_teaser.tex
\begin{figure*}[!t]
\centering
\begin{tikzpicture}[
  panel/.style={inner sep=0pt, draw=gray!55, line width=0.4pt},
  lab/.style={font=\footnotesize, align=center, text width=4.4cm},
]
\node[font=\itshape] (q) at (0,0) {``The airplane on the right''};
\node[panel, below=5pt of q, anchor=north, xshift=-6.05cm] (p1)
  {\includegraphics[width=4.4cm]{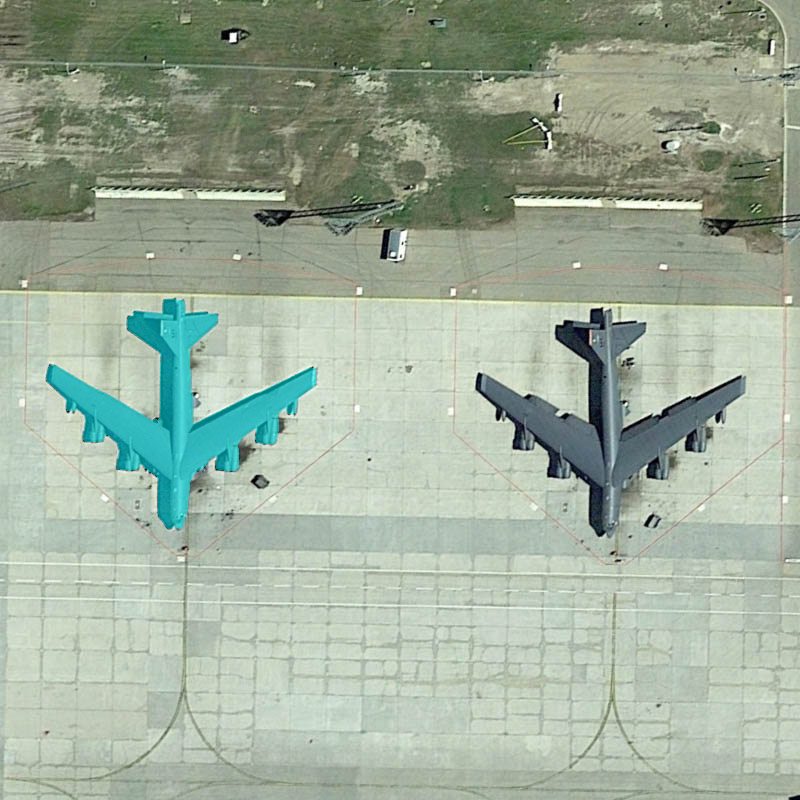}};
\node[panel, below=5pt of q, anchor=north] (p2)
  {\includegraphics[width=4.4cm]{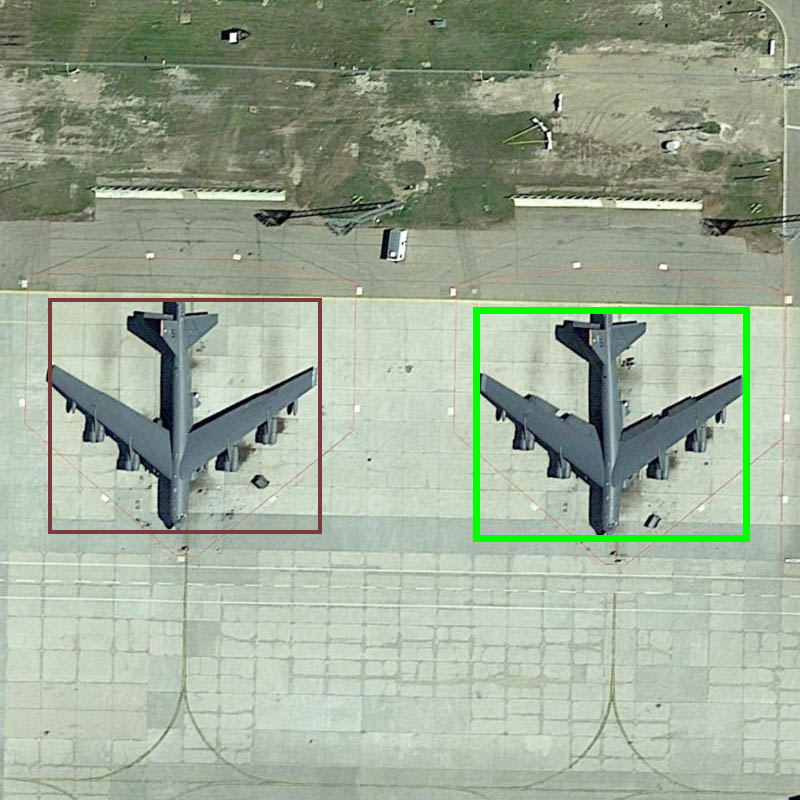}};
\node[panel, below=5pt of q, anchor=north, xshift=6.05cm] (p3)
  {\includegraphics[width=4.4cm]{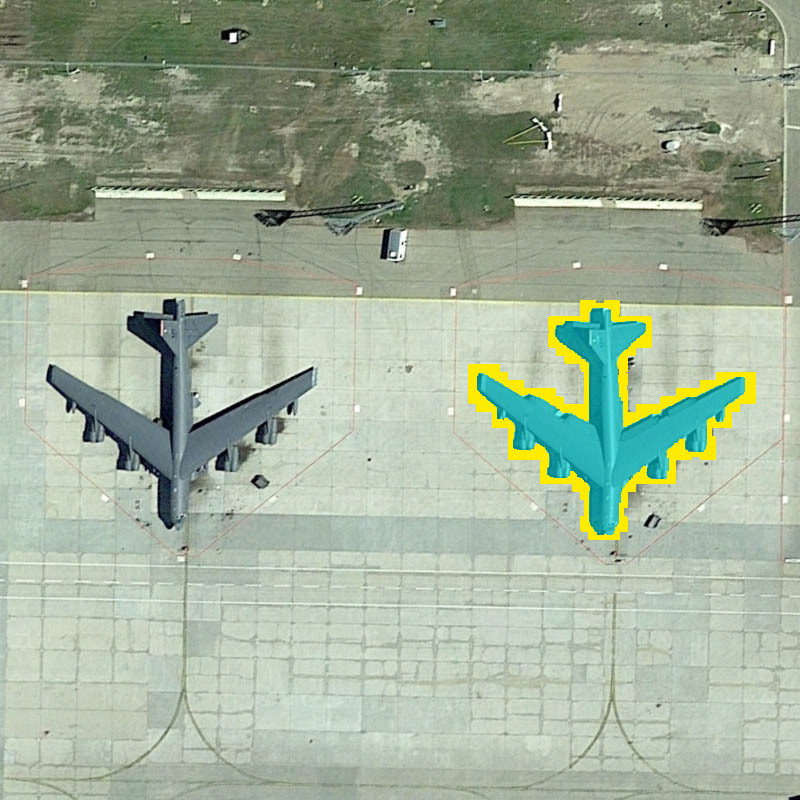}};
\node[lab, below=2pt of p1]
  {Implicit region--text matching (GeoRSCLIP):\\ wrong instance~{\color{red}\boldmath$\times$}};
\node[lab, below=2pt of p2]
  {Our executed spatial program:\\ candidates scored, the right one wins};
\node[lab, below=2pt of p3]
  {Explicit execution (GeoSelect, \textbf{ours}):\\ correct instance~{\color{green!50!black}$\checkmark$}};
\end{tikzpicture}
\caption{\textbf{Why explicit execution.} For a spatial expression over several same-class
objects, implicit region--text matching (left, GeoRSCLIP) selects the wrong instance, whereas
GeoSelect synthesises and executes a typed spatial program that scores the candidate set
(middle, brighter~$=$~higher score; the chosen box in green) and returns the instance matching the
ground truth (right). Same candidate boxes and segmenter; only the selector differs. Yellow
outline: ground-truth target.}
\label{fig:teaser}
\end{figure*}
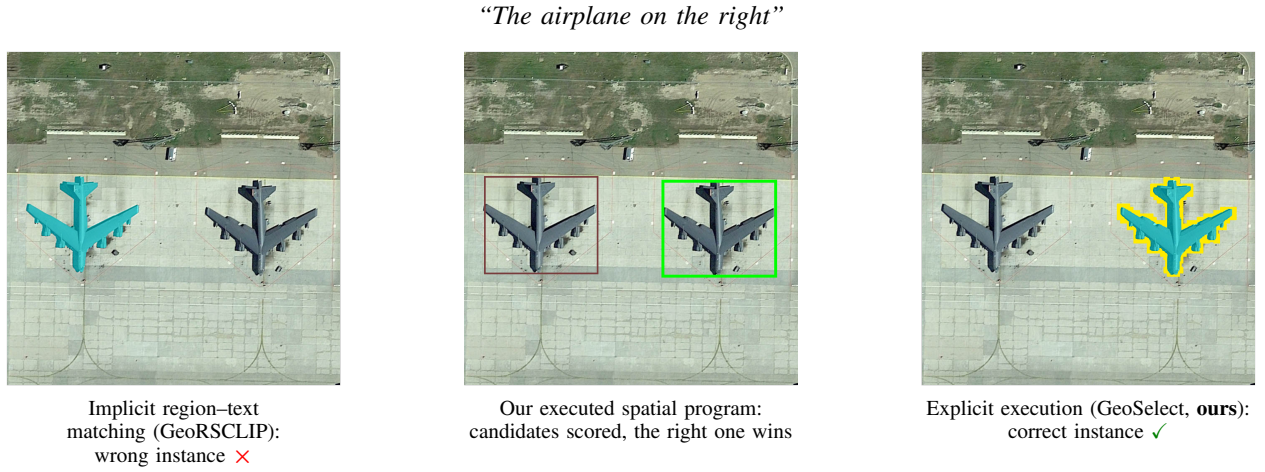

%% file: sections/related_work.tex
\section{Related Work}
\label{sec:related}

\subsection{Referring Remote Sensing Image Segmentation}
Referring image segmentation (RIS) originated in natural images, where CLIP-driven
CRIS~\cite{cris}, the transformer LAVT~\cite{lavt}, and earlier recurrent
models~\cite{rrn,brinet,cmpcplus} fuse a language embedding with dense visual features and
decode a mask. The aerial setting introduces overhead viewpoints, large scale variation, and
dense same-class clutter, which motivated dedicated benchmarks and architectures. RMSIN~\cite{rmsin}
introduced the RRSIS-D benchmark and a rotation-aware multi-scale interaction network;
CroBIM~\cite{crobim} contributed RISBench and a cross-modal bidirectional interaction module;
FIANet~\cite{fianet}, LGCE~\cite{lgce}, and BTDNet~\cite{btdnet} strengthen cross-modal
alignment and boundary quality; and RSRefSeg2~\cite{rsrefseg2} adapts vision foundation models
with SAM-based decoding. These supervised specialists reach roughly 61--69~mIoU on RRSIS-D and
serve as the learned-model reference in our experiments, but each requires dense referring
masks and a separate training run per benchmark, and most release distinct weights for RRSIS-D
and RISBench. Our setting removes both requirements and instead composes frozen perception
modules, trading learned end-to-end fusion for an explicit selection rule.

\subsection{Training-Free Referring Grounding and Segmentation}
Training-free RRSIS resolves the expression with frozen pretrained components.
EKP-HRM~\cite{ekphrm} decomposes expressions hierarchically and scores FreeSOLO masks by
GeoRSCLIP cosine similarity; RSVG-ZeroOV~\cite{rsvgzeroov} couples Qwen2.5-VL with a
diffusion grounder; DGL-RSIS~\cite{li2026dgl} reads masks from enhanced Grad-CAM activations;
and text-promptable segmenters such as SAM3~\cite{sam3} can be queried with the expression
directly. What these methods share is that the referring decision is taken implicitly,
through vision--language activations, attention, or region--text similarity, rather than by
explicitly evaluating the parsed spatial relation. The published training-free ceiling on
RRSIS-D test has remained far below the supervised specialists, which suggests that pretrained
semantic matching alone does not resolve the spatial and ordinal content of aerial referring
expressions. We adopt the same frozen-component constraint but make the relation an executable
object. Adjacent training-free remote-sensing segmenters target other tasks: SegEarth-OV~\cite{li2025segearth_ov}
and its SAM3-based successor SegEarth-OV3~\cite{li2026segearthov3} perform open-vocabulary semantic
segmentation, and InstructSAM~\cite{zheng2025instructsam} pairs a
frozen LVLM with SAM and a remote-sensing CLIP through structured output for instruction-oriented
recognition; neither evaluates the relational referring decision we study. We use the term
training-free in the strict sense of Section~\ref{sec:problem:tf}: trained RRSIS vision--language
systems~\cite{segearthr1,geoground,text4seg} are excluded even when evaluated zero-shot, and a
concurrent SAM-prompting method~\cite{trainfreeseg} achieves its strongest RRSIS-D result
($67.6$~mIoU) only by LoRA-tuning its language model; its fully training-free variant reaches
$24.9$, below GeoSelect's $58.86$.

\subsection{Neuro-Symbolic Reasoning and Visual Program Synthesis}
Resolving a compositional query by executing a program over perception modules has a long
lineage in natural images. Neural module networks~\cite{nmn} assemble a layout of learned
modules per question; NS-VQA~\cite{nsvqa} and the neuro-symbolic concept learner~\cite{nscl}
run a symbolic program over an object-centric scene representation; and prompting-era methods
such as VisProg~\cite{visprog} and ViperGPT~\cite{vipergpt} have a frozen LLM emit a program
over off-the-shelf vision APIs. For referring specifically, ReCLIP~\cite{reclip} decomposes an
expression into subject, relation, and object and scores candidate regions with frozen CLIP.
GeoSelect shares this family's spirit of a frozen LLM that synthesises a typed,
deterministically executed program, but differs in the \emph{executed representation}. VisProg
and ViperGPT dispatch to neural black-box subroutines whose spatial reasoning is opaque and
whose outputs carry no type or correctness guarantee, and ReCLIP scores relations through CLIP
similarity rather than geometry. In GeoSelect each spatial predicate is a closed-form geometric
field or set/order operator over a single scored-candidate-set type
(Eq.~\ref{eq:scored-set}), so every intermediate field and ranking is inspectable and
checked rather than dispatched to an opaque subroutine. The result is a neuro-symbolic
referring pipeline whose symbolic half is an explicit geometric calculus rather than a set of
learned modules.

\subsection{Spatial Relation Modelling, Detection, and Promptable Segmentation}
The geometric fields we execute connect to a body of work on representing spatial language as
graded, composable maps. Fuzzy spatial-relation models render directional and distance terms
as membership fields and combine them with fuzzy connectives~\cite{matsakis1999force,bloch2005fuzzy},
and cognitive spatial templates model where a relation places its referent relative to an
anchor~\cite{regier2001grounding}. Our monotone and extremal kernels are a closed-form,
benchmark-calibrated instance of this idea, connected to a text-only program
synthesiser and an aerial detector. In remote sensing, scene-graph
methods~\cite{star,recon1m,aug} predict dense relation graphs over a whole image, whereas RRSIS
grounds one queried relation; RSGround-R1~\cite{rsgroundr1} and ProVG~\cite{provg} decompose
referring expressions into reasoning steps but train their grounding models. Candidate
generation relies on open-vocabulary detection: generic detectors such as GLIP~\cite{glip} and
Grounding-DINO~\cite{gdino} transfer poorly to overhead imagery, while LAE-DINO~\cite{laedino}
pretrained on the aerial LAE-1M corpus achieves substantially stronger open-set detection on DIOR, and remote-sensing
vision--language encoders~\cite{georsclip,remoteclip,skyscript} provide semantic matching but
not precise spatial execution. The final mask comes from a promptable
segmenter~\cite{sam,ravi2025sam,sam3,yao2025remotesam}; GeoSelect treats this stage as
interchangeable and assigns the referring decision entirely to the executed program.

Remote-sensing foundation models and in-context segmenters address related prompting settings.
SkySense++~\cite{skysensepp} learns transferable multi-modal Earth-observation representations, while
SkySense-VITA~\cite{skysensevita} supports cross-image visual, textual, and fused prompting. GeoSelect
instead maps relational referring expressions to an executable spatial program.

%% file: sections/problem_formulation.tex
\section{Problem Formulation and Taxonomy}
\label{sec:problem}

This section fixes the task, the training-free setting, the selection-by-candidates
formulation that our method instantiates, and a taxonomy of referring expressions that
organises the experimental analysis. The taxonomy is defined on the language alone and is
independent of any method, so it partitions a fixed benchmark identically for every system we
compare.

\subsection{Task Definition}
\label{sec:problem:task}
Given an aerial image $I\in\mathbb{R}^{H\times W\times 3}$ and a natural-language referring
expression $E$, referring remote sensing image segmentation (RRSIS) produces a binary mask
$M\in\{0,1\}^{H\times W}$ of the single object that $E$ designates. The expression typically
names an object category together with disambiguating cues: an \emph{attribute}
(\emph{``the gray harbor''}) or a \emph{position} relative to the image frame or to another
object (\emph{``the ship in the upper right''}, \emph{``the vehicle left of the building''}).
A third cue is a \emph{set restriction} that singles out one instance from several of the same
class (\emph{``the rightmost tank''}, \emph{``the second court from the left''}). The defining
difficulty of RRSIS, relative to category segmentation, is that several same-class objects
are usually present and only the relational and ordinal content of $E$ separates the referent
from its distractors.

\subsection{Training-Free Setting}
\label{sec:problem:tf}
We call a method \emph{training-free} when no parameter of any component is updated on
referring-segmentation supervision from the target benchmark; pretrained components may retain
their original pretraining. Formally, a system is a composition of modules
$f_{\theta_1},\dots,f_{\theta_m}$ whose weights $\theta_i$ are frozen at their public
checkpoints, with no gradient taken on RRSIS-D or RISBench image--expression--mask triples.
Under this definition a vision--language model fine-tuned for referring segmentation is not
training-free even when evaluated zero-shot on a held-out split, whereas an open-vocabulary
detector that retains its detection pretraining is admissible because that pretraining is not
referring supervision. This setting removes per-benchmark mask annotation and per-benchmark
training runs, at the cost of having to define the referring decision explicitly rather
than learning it.

\subsection{Selection by Candidates}
\label{sec:problem:select}
We factor RRSIS into three training-free stages and locate the referring decision entirely in
the middle one:
\begin{equation}
\underbrace{\mathcal{B}=\mathcal{D}(I,\,Q(E))}_{\text{candidate generation}}
\;\longrightarrow\;
\underbrace{b^\star=\Pi(E,\mathcal{B},I)}_{\text{selection}}
\;\longrightarrow\;
\underbrace{M=\mathcal{S}(I,b^\star)}_{\text{segmentation}} .
\label{eq:pipeline}
\end{equation}
A frozen open-vocabulary detector $\mathcal{D}$, queried with category and attribute terms
$Q(E)$ extracted from the expression, returns a set of candidate boxes
$\mathcal{B}=\{b_1,\dots,b_N\}$; a selection operator $\Pi$ chooses the referent box
$b^\star$; and a frozen promptable segmenter $\mathcal{S}$ turns $b^\star$ into the mask.
Stages $\mathcal{D}$ and $\mathcal{S}$ are standard frozen perception modules; the scientific
content of a training-free method is the design of $\Pi$, which must evaluate the spatial,
superlative, and ordinal content of $E$ without learned parameters. Implicit methods realise
$\Pi$ as a region--text similarity (rank $b_j$ by the cosine between a vision--language
embedding of the crop $b_j$ and of $E$) or as a detector-confidence ranking; both reduce the
relation in $E$ to a single matching score. We instead realise $\Pi$ as an executable program,
described in Section~\ref{sec:method}.

\subsection{Scored Candidate Sets}
\label{sec:problem:type}
To unify continuous geometric scoring with discrete set and order operations, we type every
intermediate value as a \emph{scored candidate set}
\begin{equation}
\mathcal{C}=\big\{(b_j,\,s_j)\big\}_{j=1}^{N},\qquad b_j\in\mathbb{R}^4,\; s_j\in[0,\infty),
\label{eq:scored-set}
\end{equation}
a list of boxes each carrying a non-negative score. The detector output is the initial set
with $s_j$ the detection confidence; a continuous geometric predicate re-weights the
scores in place; a discrete operator such as \textsc{argmax} or \textsc{nth} re-ranks or
prunes the set; and conjunction and disjunction combine two sets by score algebra. Every operator
returns a scored candidate set, so continuous fields and
discrete combinators compose freely in any order, and the final referent is the top-scoring
element $b^\star=\arg\max_j s_j$. This single type is what lets one executor host both the
field algebra inherited from continuous geometry and the set/order operators that the longer
expressions require.

\subsection{Expression Taxonomy}
\label{sec:problem:taxonomy}
We group expressions by the reasoning the selector must perform, using a deterministic,
method-independent text rule applied to $E$ alone. Three categories follow prior
RRSIS practice and three extend it to the compositional constructions that motivate this work:
\begin{itemize}
\item \textbf{Image-relative}: position relative to the image frame
  (\emph{``the field in the lower left''}). The anchor is always available.
\item \textbf{Object-anchored}: position relative to another named object
  (\emph{``the car to the right of the roundabout''}), requiring the anchor to be detected and
  bound to the correct instance.
\item \textbf{Attribute}: disambiguation by an intrinsic property with no spatial relation
  (\emph{``the white storage tank''}); the selector reduces to the detector ordering.
\item \textbf{Superlative}: an extremum over a set (\emph{``the right-most ship''},
  \emph{``the top-most court''}), requiring an \textsc{argmax}/\textsc{argmin} over an axis.
\item \textbf{Ordinal}: a ranked position within a set
  (\emph{``the second vehicle from the left''}), requiring an \textsc{nth} selection.
\item \textbf{Compositional}: two or more of the above chained
  (\emph{``the building in the upper left near the river''}), requiring nested
  execution of fields and operators.
\end{itemize}
The first three are base types expressible by continuous fields alone; the last three are operator
constructions that a closed positional lexicon cannot represent and that the discrete operators of
Section~\ref{sec:method:discrete} target. Each expression carries exactly one base tag together
with any operator tags it triggers, assigned by a deterministic text rule. On a stratified sample
of $300$ validation expressions labelled independently by two annotators, the two agree perfectly on
the base frame type and on the ordinal, superlative, and compositional tags (Cohen's $\kappa=1.0$),
and substantially on the multi-instance tag ($\kappa=0.66$), which is the borderline case in these
single-instance benchmarks. The text rule reproduces these human labels on $98.0\%$ of ordinal,
$96.3\%$ of superlative, $79.0\%$ of compositional, and $87.0\%$ (annotator A) and $86.0\%$
(annotator B) of multi-instance tags. The base
frame type is a three-way label (image-relative, object-anchored, attribute) that the text rule
reproduces on $67.3\%$ of the sample; separating a frame-relative cue from an object-anchored one from
text alone is the hard case, so we treat this stratification as noisy. These tags are used only for descriptive stratification and never for prediction. The
operator strata therefore overlap the base strata and one another, so per-stratum counts need not
sum to the split size. The tags are derived from text and never from predictions, so
stratifying any method's per-sample scores by tag re-buckets fixed numbers and leaves every
overall metric unchanged.

\subsection{Evaluation Protocol}
\label{sec:problem:eval}
We follow the RMSIN protocol~\cite{rmsin}: predicted and ground-truth masks are resized to
$480\times480$ by nearest-neighbour interpolation before scoring, so that all methods are
compared at a common resolution. We report mean intersection-over-union
(mIoU, the per-sample IoU averaged over the split, equivalent to gIoU), overall IoU
(oIoU, the ratio of summed intersection to summed union, equivalent to cIoU), and precision
Pr@$X$ at IoU thresholds $X\in\{0.5,0.6,0.7,0.8,0.9\}$. mIoU weights every expression equally
and is our primary metric; oIoU is dominated by large referents and exposes under-coverage of
big objects by the single-mask decoder.

%% file: sections/method.tex
\section{Method}
\label{sec:method}

\subsection{Overview}
\label{sec:method:overview}
GeoSelect resolves a referring expression $E$ as the execution of a typed spatial program
(Fig.~\ref{fig:pipeline}; the full procedure is Algorithm~\ref{alg:executor}). A frozen, text-only language model synthesises a program
$\pi$ in a small domain-specific language (DSL) from the expression alone
(Section~\ref{sec:method:program}). A well-formedness checker accepts or rejects $\pi$. A legal program is
run by a deterministic executor that evaluates each operator over the scored candidate sets of
Eq.~\eqref{eq:scored-set}, so continuous geometric fields (Section~\ref{sec:method:fields}) and
discrete set and order operators (Section~\ref{sec:method:discrete}) run under one dispatch loop.
If a program is ill-formed, fails, or returns nothing, a reliability ladder
(Section~\ref{sec:method:reliability}) falls back to the continuous-field selector, recovering the
field-only result exactly rather than losing the answer. The top-scoring candidate is segmented by a
frozen promptable decoder. No parameter of the parser, detector, or segmenter is updated on
referring supervision. Concretely, GeoSelect instantiates the abstract selection stage $\Pi$ of
Eq.~\eqref{eq:pipeline} as the synthesise-then-execute factorisation
\begin{equation}
\begin{aligned}
\pi &= \textsc{Synthesise}(E),\\[2pt]
b^\star &= \textsc{Execute}\big(\pi,\,\textsc{Detect}(I)\big),\\[2pt]
M &= \textsc{Segment}(I,b^\star),
\end{aligned}
\label{eq:exec-pipeline}
\end{equation}
where only $\textsc{Synthesise}$ reads language and only $\textsc{Detect}$ and
$\textsc{Segment}$ read pixels, so the spatial reasoning is isolated in the symbolic executor.

Three design principles organise the method. (i)~\emph{One closed type}: every operator maps a
scored candidate set to another (Eq.~\eqref{eq:scored-set}), so continuous fields and discrete
operators compose and nest under a single type instead of separate stages.
(ii)~\emph{Closed-form geometry}: each spatial predicate is an explicit geometric computation, not
a learned or similarity-scored module, so every intermediate field and ranking is inspectable.
(iii)~\emph{A reliability floor}: a field-only special case of the same algebra is retained as a
deliberate component that any failing program falls back to, so expressive power never costs an
answer.

\input{figures/fig_pipeline}

Principle~(iii) is concrete. Our field-only special case maps each
expression to a fixed relation over positional and proximity predicates and
composes their fields by $\min/\max$. That representation covers image-relative,
object-anchored, and attribute expressions but not the extremum, rank, and top-$k$
constructions, which act on the candidate set rather than the image plane. We retain this field algebra verbatim as one
family of operators and add the missing set and order operators as a second family, so that
the longer compositional expressions of Section~\ref{sec:problem:taxonomy} are synthesised
from a common set of primitives rather than enumerated as new predicates.

\subsection{The Spatial-Program DSL}
\label{sec:method:program}
A program is a typed expression tree whose leaves generate candidates and whose internal nodes
transform scored candidate sets. Fig.~\ref{fig:grammar} gives the grammar. The leaf
\textsc{Select}$(t)$ queries the detector for noun phrase $t$ and returns the initial scored
set; \textsc{Filter} re-weights a set by a continuous geometric field;
\textsc{Argmax}/\textsc{Argmin}, \textsc{Nth}, and \textsc{RestrictCount} are the discrete
set/order operators; \textsc{Relate} scores a set by a binary spatial relation to an anchor;
and \textsc{And}/\textsc{Or}/\textsc{Not} combine sets. \textsc{Select} has type
$\textsf{Phrase}\!\rightarrow\!\mathcal{C}$, the unary operators have type
$\mathcal{C}\!\rightarrow\!\mathcal{C}$, and \textsc{And}/\textsc{Or} have type
$\mathcal{C}\!\times\!\mathcal{C}\!\rightarrow\!\mathcal{C}$, so the tree is well-formed exactly when
its leaves are \textsc{Select} nodes and operator arities match.
The operator set targets the expression space of remote-sensing referring, not a single benchmark: it
covers position, extremum, ordinal, count, relation, and logical combination. Position and extrema
dominate the two benchmarks studied here; the count, relation, and logical operators support the less
frequent compositional expressions the domain admits. Table~\ref{tab:mapping} maps a representative
phrase to each.

\paragraph*{Well-formedness checking}
Before execution, a checker verifies that (i) every candidate-producing argument ultimately
roots in a \textsc{Select} leaf; (ii) \textsc{Nth} carries a non-negative index and
\textsc{RestrictCount} a positive count; (iii) the axis and polarity of an extremum are set;
and (iv) a binary relation that requires a second anchor (\textsc{Between}) supplies one. The
checker returns the list of violated rules; a program with an empty list is \emph{legal} and
is executed, otherwise the ladder of Section~\ref{sec:method:reliability} takes over. A small
deterministic repair pass runs first and fixes the most common synthesis slips without a model
call: it drops a stray second anchor on a unary relation and promotes an anchor to the target
when the nominal target is the image frame. Well-formedness checking and repair are model-free and cost
microseconds.

\paragraph*{Semantics}
The program denotes a function from the image to a single box. Leaves are evaluated against the
detector output; each internal node consumes the scored set(s) of its children and emits a
scored set as defined below; and the referent is the top-scoring element of the root set,
$b^\star=\arg\max_j s_j$. The type is closed under every operator, so fields and discrete
operators interleave in any order, which is what allows
\textsc{Argmax}$(\textsc{Filter}(\textsc{Select}(t),\,\phi))$ for a directional predicate $\phi$,
``the rightmost among those in the upper left'', to be built from the same primitives as a bare
directional filter.

\input{figures/fig_grammar}

\subsection{Continuous Geometric Fields}
\label{sec:method:fields}
The \textsc{Filter} operator re-weights a candidate set by a dense geometric field
$P\in[0,1]^{H\times W}$ evaluated at each candidate. For positional predicates we reuse the
closed-form, anchor-relative kernels of the field-only special case, which we summarise here
for completeness; they are the \emph{monotone} kernel family of
Section~\ref{sec:method:kernels}. Let $(c_x,c_y)$ be the centre of anchor box $A$. For a
cardinal direction $d$,
\begin{equation}
\begin{aligned}
q_d(x,y)&=
\begin{cases}
x-c_x, & d=\textsc{right},\\
c_x-x, & d=\textsc{left},\\
y-c_y, & d=\textsc{bottom},\\
c_y-y, & d=\textsc{top},
\end{cases}\\
P_d(x,y)&=\left[\frac{\operatorname{clip}(q_d(x,y),0,L_d)}{L_d}\right]^\gamma,
\end{aligned}
\label{eq:direction}
\end{equation}
with $L_d=W$ for horizontal and $L_d=H$ for vertical directions and $\gamma=1$ in the reported
configuration. A diagonal field is the pixelwise minimum of its two cardinal fields. Scores are read at
candidate box centres, which lie within the image domain
$[0,W]\times[0,H]$, so the clip never has to separate two valid candidates; it bounds the
normalisation and does not change their ordering. The
centre and proximity kernels are
\begin{equation}
\begin{aligned}
P_{\textsc{center}}(x,y)&=\exp\!\Big[-\tfrac{(x-c_x)^2+(y-c_y)^2}{2\sigma_c^2}\Big],\\
P_{\textsc{near}}(x,y)&=\exp\!\Big[-\tfrac{d_A(x,y)^2}{2L_A^2}\Big],
\end{aligned}
\label{eq:center-near}
\end{equation}
with $d_A(x,y)$ the Euclidean distance to the nearest point of $A$ (zero inside the box),
$\sigma_c=0.25\min(H,W)$, and $L_A=1.5(w_A+h_A)/2$. A candidate box $b_j$ with centre
$(c_x^j,c_y^j)$ is scored by the field at its centre,
\begin{equation}
\rho(b_j)=P\big(c_x^j,\,c_y^j\big),
\label{eq:boxscore}
\end{equation}
and \textsc{Filter} updates its score multiplicatively, $s_j\!\leftarrow\!s_j\,\rho(b_j)$, so
that a chain of filters realises a conjunction of constraints. Centre sampling makes the score
independent of box size, which avoids the systematic dilution that a region mean inflicts on
large referents (a big central object would otherwise score below a tiny box sitting on the
prior peak). It also yields a useful invariance: because $\rho$ reads a single point, the
selection $\arg\max_j s_j$ is unchanged under any strictly increasing reshaping of a field.
The directional exponent $\gamma$ and the widths $\sigma_c,L_A$ alter field magnitudes but not
the per-box ranking, so the headline is exactly invariant to them
(Section~\ref{sec:exp:param}). These shape parameters thus drop out of the tuning surface; only
the candidate set and the scoring rule remain as design choices. For an image-relative predicate
the anchor is the image frame; for an object-relative predicate the anchor is resolved by
recursively executing the anchor sub-program and taking its top-scoring box, which lets an
anchor itself be constrained (\emph{``right of the building in the upper left''}).
\textsc{And}, \textsc{Or}, and \textsc{Not} combine scored sets matched by box identity,
\begin{equation}
\begin{aligned}
(\mathcal{C}_1\wedge\mathcal{C}_2)_j &= \min(s^1_j,s^2_j),\\[2pt]
(\mathcal{C}_1\vee\mathcal{C}_2)_j &= \max(s^1_j,s^2_j),\\[2pt]
(\neg\,\mathcal{C})_j &= 1-s_j,
\end{aligned}
\label{eq:setalg}
\end{equation}
a non-compensatory conjunction in which a high response to one predicate cannot offset the
violation of another; under centre scoring this $\min/\max$ on scores equals the pixelwise
$\min/\max$ of the underlying fields, preserving the field-only equivalence. The complement
$1-s_j$ is composed under $\wedge$ with a positive constraint, so it acts as a soft
exclusion mask that down-weights the excluded candidates and never stands alone as a selector.
Fig.~\ref{fig:prior} visualises the field and the grounding it produces for representative
image-relative predicates.
\input{figures/fig_prior} Restricting
the program to \textsc{Select}, \textsc{Filter}, \textsc{And}, and \textsc{Or} with the
monotone kernel and the centre score of Eq.~\eqref{eq:boxscore} reproduces the field-only
special case exactly, which fixes it as the reliability floor.

\subsection{Discrete Compositional Operators}
\label{sec:method:discrete}
The set and order operators are the additions that cover the superlative, ordinal, and
compositional expressions of the taxonomy, including a top-$k$ restriction operator. Each is a
closed-form map on the scored set; none introduces a learned parameter.

\paragraph*{Extremum (\textsc{Argmax}/\textsc{Argmin})}
A positional superlative selects the extreme candidate along an axis $u$, the signed projection of its
centre onto a direction (image $x$, image $y$, or an image diagonal). Hard selection is brittle under detector noise,
so we instead \emph{replace} the scores of the live set by a temperature-controlled softmax over
$u$. The incoming score defines membership in the live set rather than acting as a weight:
candidates an earlier filter drove to zero are excluded, and if none survive the whole set is
live. For $j$ in the live set $L$,
\begin{equation}
s_j \;\leftarrow\;
\frac{\exp\!\big(\epsilon\,\beta\,u(b_j)\big)}{\sum_{k\in L}\exp\!\big(\epsilon\,\beta\,u(b_k)\big)},
\qquad \epsilon\in\{+1,-1\},
\label{eq:extremal}
\end{equation}
and $s_j\!=\!0$ for $j\notin L$, with $\epsilon=+1$ for \textsc{Argmax} and $-1$ for
\textsc{Argmin}. The temperature is set to $\beta=1/\sigma_u$, the inverse standard deviation of
the candidates' axis values $\{u(b_j)\}_{j\in L}$, so the softmax exponent is measured in
spread units and the operator is invariant to the absolute scale of the scene; a degenerate set
($\sigma_u\!=\!0$, all candidates tied on the axis) gives uniform scores. Replacing rather than scaling is what stops a superlative from
penalising a candidate for its incoming score: the rightmost of a set of left-side ships is not
docked for being rightmost. The result stays a graded set that a downstream operator can read.
The axis polarity is fixed by the parser: \emph{top-most} maps to \textsc{Argmin} on image $y$,
and \emph{right-most} to \textsc{Argmax} on image $x$. The reported configuration supports these
positional superlatives, namely leftmost, rightmost, topmost, and bottommost. Metric-size
superlatives and comparatives, such as \emph{largest}, \emph{smaller than}, and \emph{similar in
size}, are outside the claimed coverage. A non-comparative size adjective may remain in the
detector-query noun phrase, but the system does not rank instances by size
(see the Limitations in Section~\ref{sec:discussion}).

\paragraph*{Ordinal (\textsc{Nth})}
An ordinal expression (\emph{``the second tank from the left''}) ranks the live candidates in
reading order along the axis $u$ and keeps the $n$-th, with $n$ a zero-based index supplied by
the parser:
\begin{equation}
s_j \;\leftarrow\; s_j \cdot \mathbb{1}\!\big[\,\operatorname{rank}_u(b_j)=n\,\big],
\label{eq:nth}
\end{equation}
with $\operatorname{rank}_u$ ascending in $u(b_j)$, so the kept box retains its incoming score
and the rest are zeroed; the operator prunes the set to one element while remaining
type-compatible.

\paragraph*{Top-$k$ candidate restriction (\textsc{RestrictCount})}
A count in the expression (\emph{``the rightmost of the three ships''}) restricts
the scored set to its top-$k$ candidates by current score, with $k$ from the parser:
\begin{equation}
s_j \;\leftarrow\; s_j \cdot \mathbb{1}\!\big[\,j \in \operatorname{top\text{-}}k(s)\,\big].
\label{eq:topk}
\end{equation}
\textsc{RestrictCount} does not return a union of masks. It narrows the candidate set that a
subsequent operator, typically \textsc{Argmax}/\textsc{Argmin} or \textsc{Nth}, selects from; when
no count is given it is the identity. The single-instance evaluation here scores one mask per
expression; multi-mask output is controlled separately by the parser's group cardinality, which fires
on a small number of group-tagged RISBench expressions.

\paragraph*{Binary relations (\textsc{Relate})}
Beyond frame- and centre-relative position, \textsc{Relate} scores candidates by a binary
spatial relation to an anchor set: \textsc{Near}, \textsc{Between} (two anchors),
\textsc{Adjacent}, \textsc{Contains}, and \textsc{Inside}. Each relation is a closed-form score
$r(b_j,A)$ on box geometry, \emph{max-marginalised} over anchor instances and multiplied into
the candidate score:
\begin{equation}
\begin{gathered}
s_j \;\leftarrow\; s_j\cdot\max_{A\in\mathcal{A}} r(b_j,A),\\[4pt]
r_{\textsc{inside}}(b,A) = \frac{|b\cap A|}{|b|},\\[4pt]
r_{\textsc{between}}(b,A_1,A_2) = \exp\!\Big[-\tfrac{d(c_b,\overline{c_{A_1}c_{A_2}})^2}{2\sigma^2}\Big],
\end{gathered}
\label{eq:relate}
\end{equation}
where $c$ denotes a box centre and $d(c_b,\overline{c_{A_1}c_{A_2}})$ is the perpendicular
distance from the candidate centre to the segment joining the two anchor centres (so
\textsc{Between} peaks on that segment). The \textsc{Between} score is additionally gated by a
betweenness factor that decays once the candidate projects past either endpoint of the segment,
and its width scales with the geometry, $\sigma=0.35\lvert A_1A_2\rvert$ plus the candidate's
characteristic length, so the tolerance grows with anchor separation and target size.
\textsc{Contains} is $|b\cap A|/|A|$ and \textsc{Near}/\textsc{Adjacent} use the proximity kernel
of Eq.~\eqref{eq:center-near}. Max-marginalisation means a relation to ``a building'' is satisfied
by the best-placed building.

\subsection{Dual Geometric Kernels}
\label{sec:method:kernels}
The operators above draw their pixel- or axis-level scores from two kernel families with
complementary shapes. The \emph{monotone} kernel (Eqs.~\ref{eq:direction}--\ref{eq:center-near})
increases without bound away from the anchor along the named direction and is correct for
\emph{furthest-in-direction} readings (\emph{``the ship on the far right''}) and for the
field-equivalent special case. The \emph{extremal} kernel (Eq.~\ref{eq:extremal}) is a
softmax over the candidate set rather than over the image plane, and is correct for
selecting one element by an axis extremum. Separating the kernels makes explicit that
``left of'' (monotone) and ``left-most'' (extremal over the set) are different operations on
the same axis, a distinction that a single directional ramp conflates and that we trace as a
failure mode of the field-only special case (the ``Failure Modes by Execution Route'' analysis,
Section~\ref{sec:exp:failure}).

\subsection{The Executor}
\label{sec:method:exec}
A legal program is evaluated by the recursive \textsc{Exec} procedure of
Algorithm~\ref{alg:executor}. The executor walks the tree bottom-up: each
\textsc{Select} leaf is bound to the detector's candidate set for its phrase; each operator
node dispatches on its type to the corresponding map of
Sections~\ref{sec:method:fields}--\ref{sec:method:discrete}; and \textsc{And}/\textsc{Or}/%
\textsc{Not} combine child sets by the field algebra. \textsc{And}/\textsc{Or}/\textsc{Not}
match candidates by proposal identity, so their operands use the same normalised \textsc{Select} query
and its cached detections and the candidates align; reliable alignment across two different detector
queries is outside the current scope. Anchor resolution follows one rule: a
bare \textsc{Select} anchor is \emph{marginalised} (its field is the disjunction over all its
instances), whereas a \emph{constrained} anchor sub-program is resolved to its single
top-scoring box, so that ``near a building'' considers all buildings while ``near the rightmost
building'' commits to one. The procedure is deterministic, allocates one $H\times W$ field per
active filter, and runs in tens of milliseconds; its output is the ranked root set whose
maximum is $b^\star$.

\input{figures/fig_executor}

\subsection{Program Synthesis}
\label{sec:method:synth}
The synthesiser is a frozen, text-only instruction LLM (Qwen3-4B~\cite{qwen3} by default) that
reads $E$ and emits the program as JSON. The model does not see the image, so the program is
compiled from language alone and perception enters only at the detector, executor, and
segmenter. The prompt is a fixed, handcrafted few-shot schema, programmatically verified
disjoint from the validation and test expressions; ``thinking'' decoding is disabled so the
JSON is the first output. The schema encodes the design rules that a single pass otherwise
gets wrong: positional superlatives map to the extremal operator with the correct axis polarity, image
quadrants use the image frame as anchor, ``middle'' maps to \textsc{Center} (never to an
extremum), and object-relative directional cues use the monotone kernel instead of a band.
Synthesis is batched with left-padded greedy decoding, which is bit-identical to single-example
decoding and amortises the model load. Synthesis, checking, and execution are
separated, so the parser can be swapped for a different LLM without touching the executor
(Section~\ref{sec:exp:parser}).

\subsection{Reliability Ladder}
\label{sec:method:reliability}
Per-expression program synthesis trades the closed lexicon's guaranteed validity for
expressive power, so we make the failure modes explicit and bounded. The ladder is:
synthesise $\pi$; repair and check; if $\pi$ is legal, execute it; if execution raises or
returns an empty set, or if $\pi$ is ill-formed, fall back to the field-only selector on the
same candidates. The field-only selector is itself the monotone special case of
Section~\ref{sec:method:fields}, so its result is recovered exactly on every fallback; a
well-typed program may still under-perform it on an individual sample, but in aggregate the
full system does not fall below it on either benchmark (Section~\ref{sec:exp:reliability}). We
report the fraction of expressions
resolved by an executed program with a non-empty result (the \emph{program rate}) alongside accuracy,
and measure the two separately because program success is not correctness (Section~\ref{sec:exp}). Every
intermediate value (program, fields, scored sets, selected box) is inspectable, which
makes the failure analysis of Section~\ref{sec:exp:failure} possible.

\subsection{Attribute-Conditioned Detection}
\label{sec:method:detect}
Candidates come from an open-vocabulary detector (LAE-DINO~\cite{laedino}), queried for each
\textsc{Select} phrase in attribute-conditioned form. The synthesis schema keeps colour, shape, and
non-comparative size adjectives inside the noun phrase (e.g.\ \emph{``gray harbor''}) and routes
supported positional cues into operators; a deterministic stabiliser then strips positional tails from
detector queries and discards comparative-size tails, which are outside the reported scope. We run sliced inference with SAHI~\cite{sahi} at a 768-px tile size
together with a full-image pass, use a detection threshold of $0.15$, and apply per-class
non-maximum suppression; boxes are mapped to original image coordinates before execution.
Detection is cached per image and query so that it is computed once and reused across program
variants, since the candidate set depends only on the phrase and detector parameters and not
on the rest of the program.

\subsection{Segmentation}
\label{sec:method:sam}
The selected box $b^\star$ prompts the Segment Anything ViT-L decoder~\cite{sam} in single-mask
mode, and the mask is scored under the protocol of Section~\ref{sec:problem:eval}. The decoder
is frozen and interchangeable; Section~\ref{sec:exp:decoder} shows the headline moves within a
small band across SAM variants, confirming that the referring decision, not the segmenter,
carries the result.

%% file: figures/fig_pipeline.tex
\begin{figure*}[!t]
\centering
\resizebox{\textwidth}{!}{%
\begin{tikzpicture}[
  font=\footnotesize,
  stage/.style={draw, rounded corners=2pt, align=center, minimum height=0.95cm,
                text width=2.15cm, inner sep=2pt, fill=blue!5},
  hi/.style={fill=teal!12},
  art/.style={draw=gray!55, line width=0.3pt, inner sep=0pt},
  capt/.style={font=\scriptsize},
  ar/.style={-{Stealth[length=2mm]}, semithick},
  flab/.style={font=\scriptsize\itshape, midway, above=0pt},
]
\node[stage]                         (s1) {\textbf{Synthesise}\\\scriptsize text-only LLM\\(Qwen3-4B)};
\node[stage, right=1.15cm of s1]     (s2) {\textbf{\textsc{select}}\\\scriptsize detect candidates\\(``storage tank'')};
\node[stage, hi, right=1.15cm of s2] (s3) {\textbf{\textsc{filter}}\\\scriptsize continuous field\\(\textsc{left})};
\node[stage, hi, right=1.15cm of s3] (s4) {\textbf{\textsc{argmin}}\\\scriptsize discrete pick\\(axis $X$)};
\node[stage, right=1.15cm of s4]     (s5) {\textbf{Segment}\\\scriptsize SAM ViT-L};
\node[art, left=0.7cm of s1] (in) {\includegraphics[width=1.85cm]{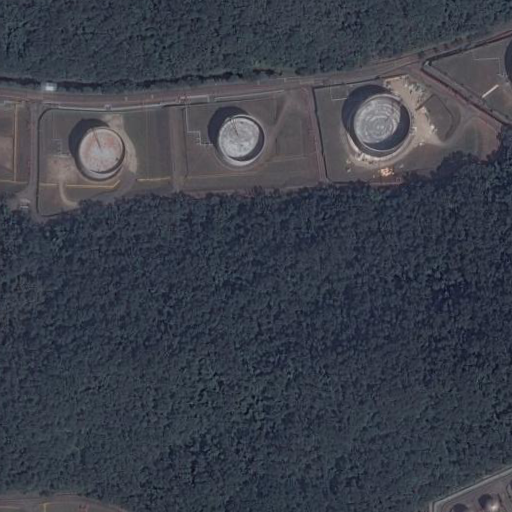}};
\node[font=\itshape, align=center, above=3pt of in, text width=2.2cm]
  {``the westernmost storage tank in the line of tanks''};
\draw[ar] (in) -- (s1);
\draw[ar] (s1) -- node[flab]{$\pi$}             (s2);
\draw[ar] (s2) -- node[flab]{$\mathcal{C}$}     (s3);
\draw[ar] (s3) -- node[flab]{$\mathcal{C}$}     (s4);
\draw[ar] (s4) -- node[flab]{box}               (s5);
\node[draw=gray!55, rounded corners=1pt, align=left, font=\scriptsize, inner sep=3pt,
      text width=2.15cm] (a1) at ($(s1)+(0,2.55cm)$)
  {\textbf{program $\pi$}\\[2pt]\textsc{argmin}(\\\hspace{0.7em}\textsc{filter}(\\\hspace{1.4em}\textsc{select},\,\textsc{left}),\\\hspace{0.7em}$X$\,)};
\node[art] (a2) at ($(s2)+(0,2.55cm)$) {\includegraphics[width=2.2cm]{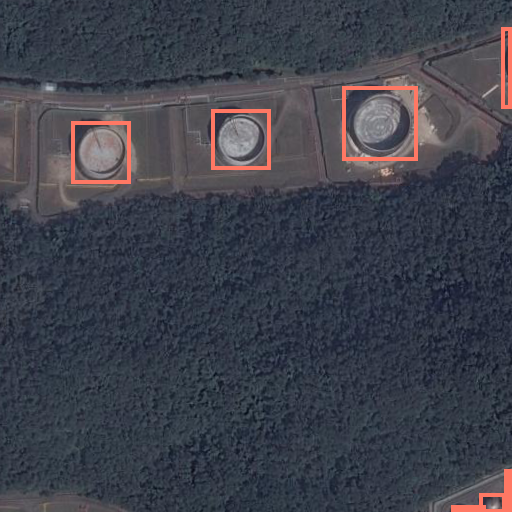}};
\node[art] (a3) at ($(s3)+(0,2.55cm)$) {\includegraphics[width=2.2cm]{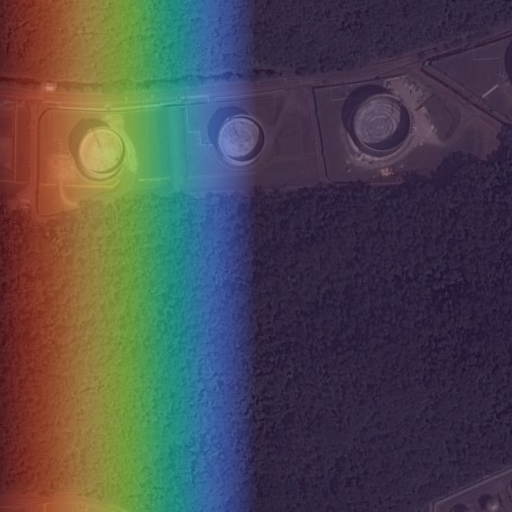}};
\node[art] (a4) at ($(s4)+(0,2.55cm)$) {\includegraphics[width=2.2cm]{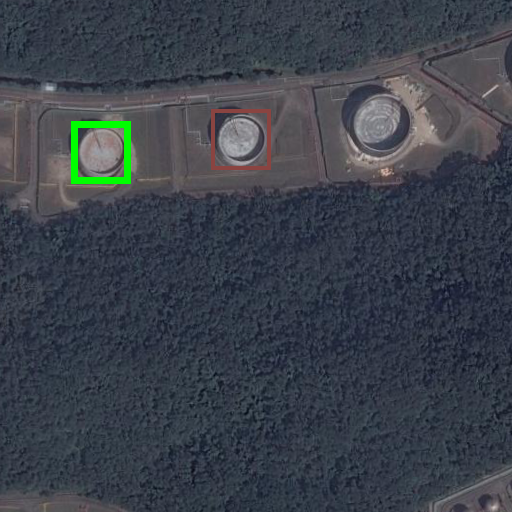}};
\node[art] (a5) at ($(s5)+(0,2.55cm)$) {\includegraphics[width=2.2cm]{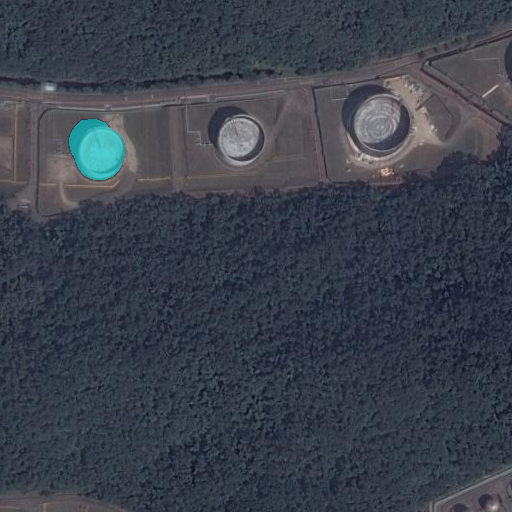}};
\node[capt, above=2pt of a2] {candidates};
\node[capt, above=2pt of a3] {geometric field $G$};
\node[capt, above=2pt of a4] {extremum (green)};
\node[capt, above=2pt of a5] {referent mask};
\node[font=\scriptsize, align=center] at ($(s3.south)!0.5!(s4.south)+(0,-0.62)$)
     {one scored candidate set $\mathcal{C}=\mathrm{List}[(\mathrm{box},\mathrm{score})]$; every operator has type $\mathcal{C}\!\rightarrow\!\mathcal{C}$};
\end{tikzpicture}}
\caption{\textbf{GeoSelect executes a typed spatial program over a single scored candidate set,
with a \emph{real} per-operator trace above each stage} (RISBench, ``the westernmost storage tank in
the line of tanks''). A frozen text-only LLM synthesises the program $\pi$; the executor evaluates it
bottom-up: \textsc{select} returns the detected candidates, the continuous \textsc{filter} (teal)
re-weights them by a geometric field, and the discrete \textsc{argmin} (teal) re-scores the set to its
extremum (green), which SAM segments.}
\label{fig:pipeline}
\end{figure*}

%% file: figures/fig_grammar.tex
\begin{figure}[t]
\centering
\footnotesize
\begin{tabular}{@{}l@{}}
\hline
\rule{0pt}{2.2ex}%
\texttt{prog\ \ \ ::= expr}\\
\texttt{expr\ \ \ ::= Select(phrase)}\\
\texttt{\ \ \ \ \ \ \ |\ Filter(expr,\ pred,\ anchor)}\\
\texttt{\ \ \ \ \ \ \ |\ Argmax(expr,\ axis)\ |\ Argmin(expr,\ axis)}\\
\texttt{\ \ \ \ \ \ \ |\ Nth(expr,\ n,\ axis)}\\
\texttt{\ \ \ \ \ \ \ |\ RestrictCount(expr,\ k)}\\
\texttt{\ \ \ \ \ \ \ |\ Relate(expr,\ rel,\ anchor\ [,\ anchor2])}\\
\texttt{\ \ \ \ \ \ \ |\ And(expr,\ expr)\ |\ Or(expr,\ expr)}\\
\texttt{\ \ \ \ \ \ \ |\ Not(expr)}\\
\texttt{anchor\ ::= IMAGE\ |\ expr}\\
\texttt{pred\ \ \ ::= LEFT|RIGHT|TOP|BOTTOM|UL|UR|LL|LR}\\
\texttt{\ \ \ \ \ \ \ \ \ \ \ |\ CENTER|NEAR}\\
\texttt{rel\ \ \ \ ::= NEAR|BETWEEN|ADJACENT|CONTAINS|INSIDE}\\
\texttt{axis\ \ \ ::= X\ |\ Y\ |\ DIAG\_UL\_LR\ |\ DIAG\_UR\_LL}\\[0.4ex]
\hline
\end{tabular}
\caption{The spatial-program DSL. \texttt{Select} produces the initial scored candidate
set~$\mathcal{C}$ from the detector and the remaining operators consume and produce candidate sets;
operator roles and type signatures are given in Section~\ref{sec:method:program}, and the complete
EBNF with production types in Appendix~\ref{sec:appendix:schema}.}
\label{fig:grammar}
\end{figure}

%% file: figures/fig_prior.tex
\begin{figure}[!t]
\centering
\includegraphics[width=\columnwidth]{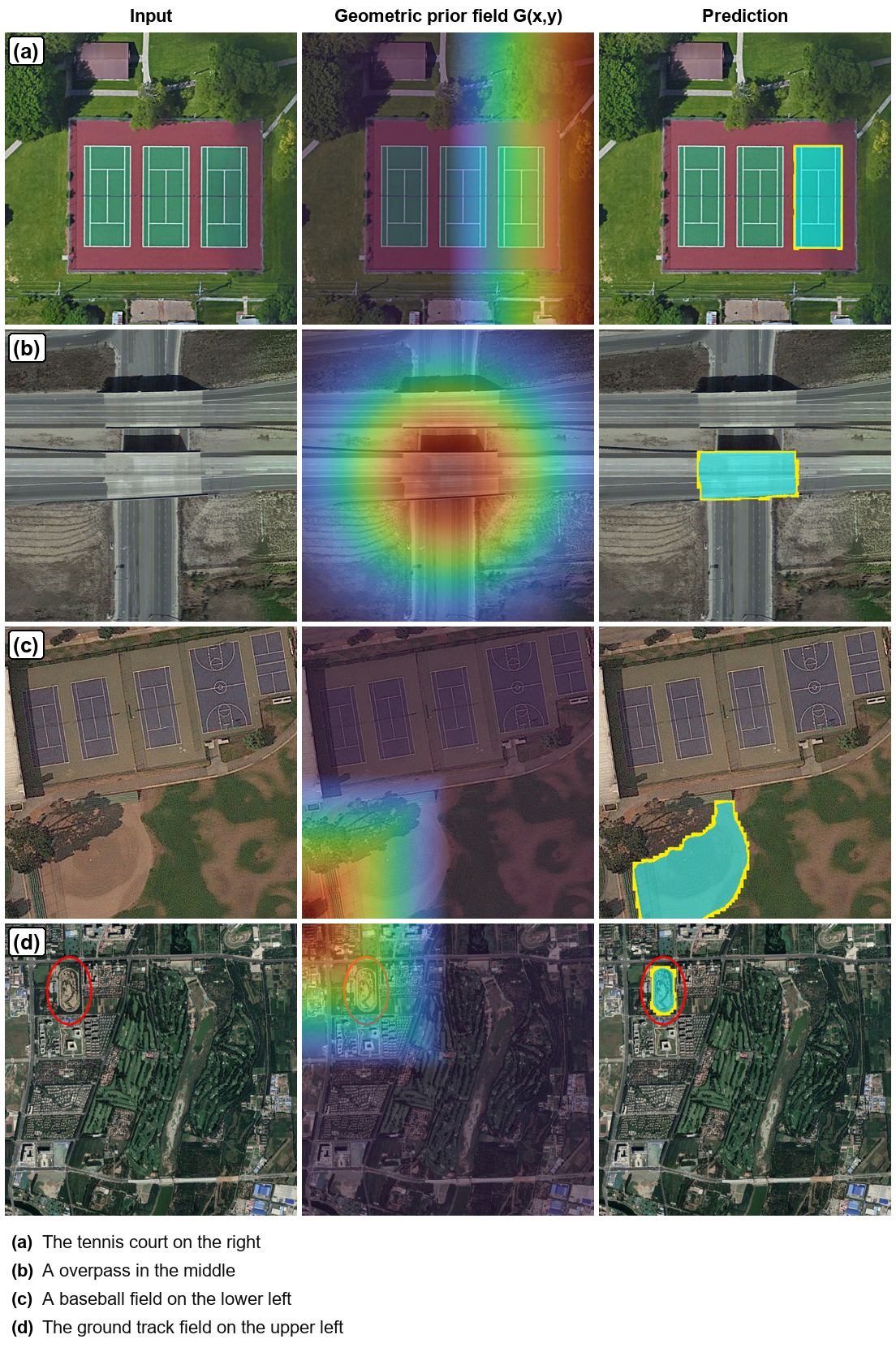}
\caption{\textbf{The continuous geometric field $G(x,y)$ of the \textsc{Filter} operator.} For
image-relative predicates the field is computed in closed form against the image frame
(Eqs.~\ref{eq:direction}--\ref{eq:center-near}); a candidate is scored at its centre
(Eq.~\ref{eq:boxscore}). Input (left), field as a heat map (middle, warm~$=$~high), and the
resulting prediction (right; cyan~$=$~mask, yellow~$=$~ground truth).}
\label{fig:prior}
\end{figure}

%% file: figures/fig_executor.tex
\begin{algorithm}[!t]
\caption{GeoSelect: synthesise, check, execute, fall back}
\label{alg:executor}
\footnotesize
\algrenewcommand{\algorithmiccomment}[1]{\hfill\textit{#1}}
\begin{algorithmic}[1]
\Require expression $E$, image $I$; frozen LLM, detector \textsc{Detect}, segmenter \textsc{Segment}
\Ensure referent mask $M$
\State $\pi\gets\textsc{Synthesise}(E)$ \Comment{text-only LLM; the image is unseen}
\State $\pi\gets\textsc{Repair}(\pi)$ \Comment{fix benign synthesis slips, model-free}
\If{\textsc{WellFormed}$(\pi)$}
   \State $\mathcal{C}\gets$ \Call{Exec}{$\pi$} \Comment{run the typed program}
   \If{$\mathcal{C}=\varnothing$ \textbf{ or } execution errs}
      \State $\mathcal{C}\gets\textsc{FieldOnly}(E,I)$ \Comment{fall back: execution failed}
   \EndIf
\Else
   \State $\mathcal{C}\gets\textsc{FieldOnly}(E,I)$ \Comment{fall back: ill-formed program (Sec.~\ref{sec:method:reliability})}
\EndIf
\State $b^\star\gets\arg\max_{b_j\in\mathcal{C}} s_j$
\State \Return $M\gets\textsc{Segment}(I,b^\star)$
\Statex
\Function{Exec}{node}
  \If{node is $\textsc{Select}(t)$}
     \State \Return $\textsc{Detect}(I,t)$ \Comment{initial candidate set $\mathcal{C}$}
  \ElsIf{node is $\textsc{And}/\textsc{Or}/\textsc{Not}$}
     \State \Return $\min/\max/\mathrm{comp}$ of children \Comment{field algebra, Eq.~\eqref{eq:setalg}}
  \EndIf
  \State $\mathcal{C}\gets$ \Call{Exec}{node.child}
  \If{node is $\textsc{Filter}(\phi,A)$}
     \State $\mathbf{A}\gets$ \Call{ResolveAnchor}{$A$};\ \ $P\gets\textsc{Field}(\phi,\mathbf{A})$
     \State $s_j\gets s_j\,P(\mathrm{centre}\,b_j)\ \ \forall b_j$ \Comment{Eq.~\eqref{eq:boxscore}}
  \ElsIf{node is $\textsc{Argmax}/\textsc{Argmin}(u)$}
     \State extremal-softmax rescore on axis $u$ \Comment{superlative, Eq.~\eqref{eq:extremal}}
  \ElsIf{node is $\textsc{Nth}(n,u)$}
     \State rank on $u$, keep the $n$-th \Comment{ordinal, Eq.~\eqref{eq:nth}}
  \ElsIf{node is $\textsc{RestrictCount}(k)$}
     \State keep top-$k$ candidates \Comment{top-$k$ candidate restriction, Eq.~\eqref{eq:topk}}
  \ElsIf{node is $\textsc{Relate}(r,A)$}
     \State $\mathbf{A}\gets$ \Call{ResolveAnchor}{$A$};\ \ $s_j\gets s_j\max_{a\in\mathbf{A}} r(b_j,a)$ \Comment{relation, Eq.~\eqref{eq:relate}}
  \EndIf
  \State \Return $\mathcal{C}$
\EndFunction
\Statex
\Function{ResolveAnchor}{$a$}
  \If{$a=\textsc{Image}$}
     \State \Return image frame
  \ElsIf{$a$ is a bare \textsc{Select}}
     \State \Return all detected instances \Comment{marginalise}
  \Else
     \State \Return top-scoring box of \Call{Exec}{$a$} \Comment{constrained anchor}
  \EndIf
\EndFunction
\end{algorithmic}
\end{algorithm}

%% file: sections/experiments.tex
\section{Experiments}
\label{sec:exp}

\subsection{Setup}
\label{sec:exp:setup}
We evaluate on RRSIS-D~\cite{rmsin}, the standard RRSIS benchmark and single-instance by
construction, and on RISBench~\cite{crobim}, which is larger and has longer, more constrained
expressions; both provide expression-conditioned pixel masks (Table~\ref{tab:datasets}).
MEET~\cite{meet} contains $1.03$ million zoom-free images for fine-grained geospatial scene
classification, but its protocol provides scene labels rather than expression-conditioned masks and
therefore does not support the mask-based evaluation here. RRSIS-D provides 17,402 image--expression pairs over 20 categories with a
fixed train/val/test split; we use the 1,740-expression val split for all design choices and
the 3,481-expression test split for the headline. RISBench is an order of magnitude larger,
with 16,159 test expressions whose distribution is far richer in superlative, ordinal,
and compositional constructions (Section~\ref{sec:problem:taxonomy}); it is used purely to measure
transfer of the RRSIS-D-tuned configuration. This two-benchmark protocol follows recent
work evaluated on both benchmarks~\cite{rsvgzeroov,provg}. We report mIoU, oIoU, and
Pr@$X$ under the RMSIN protocol of Section~\ref{sec:problem:eval} (eval@480 nearest).

The frozen configuration is a text-only Qwen3-4B synthesiser; LAE-DINO with SAHI tiling at
768~px and a detection threshold of $0.15$; attribute-conditioned queries; the typed executor
with the monotone and extremal kernels; and the Segment Anything ViT-L mask decoder in
single-mask mode, with no learned reranker. All design choices, including the mask decoder and the few-shot schema, were made on
RRSIS-D val; the configuration is frozen before any test evaluation, and no test-set recalibration
is performed on either benchmark. The handcrafted few-shot prompt is
programmatically verified disjoint from the val and test expressions. All inference runs on a
single NVIDIA GeForce RTX~3090. Results for prior methods are taken from their source papers or published
compilations~\cite{provg,crobim,rsrefseg2}; we evaluate SAM3 ourselves.
\input{tables/tab_datasets}

\input{tables/tab_main}

GeoSelect achieves 58.86~mIoU / 59.48~oIoU on RRSIS-D test (Table~\ref{tab:main};
mIoU $95\%$ bootstrap CI $[57.71,60.03]$):
$2.08$ times the best prior training-free method (RSVG-ZeroOV~\cite{rsvgzeroov}) in mIoU and
$2.6$ times in oIoU, three times the decomposition-based EKP-HRM~\cite{ekphrm}, and about
$93\%$ of the fully-supervised specialist RMSIN~\cite{rmsin} in mIoU. Its oIoU and mIoU are
close, in contrast to the supervised specialists whose oIoU
exceeds mIoU by 10--15 points; this reflects under-coverage of the largest referents by the
single-mask decoder (Section~\ref{sec:exp:decoder}). Disabling the discrete operators (the
field-only special case) gives 58.44, so on RRSIS-D, whose test split is almost entirely
image-relative, object-anchored, and attribute expressions, the operators add little; their
effect appears on RISBench (Section~\ref{sec:exp:ops}) and on the object-anchored and
compositional val strata (Table~\ref{tab:c1geom}). At stricter thresholds GeoSelect reaches
Pr@$0.5$/$0.6$/$0.7$/$0.8$/$0.9$ of $68.9$/$61.6$/$51.2$/$39.1$/$22.6$ on RRSIS-D test and
$60.3$/$56.0$/$50.7$/$44.5$/$34.7$ on RISBench test. Fig.~\ref{fig:rrsisd_qual} shows qualitative predictions across strata, and
Fig.~\ref{fig:quad} compares GeoSelect with the training-free RSVG-ZeroOV and SAM3 and the
supervised RSRefSeg2: the training-free baselines over-segment multiple same-class regions or
select the wrong instance, whereas GeoSelect resolves the relation to the intended instance
comparably to the supervised RSRefSeg2.
GeoSelect updates no parameters on either benchmark (Table~\ref{tab:efficiency}) and runs on a
single NVIDIA GeForce RTX~3090. At batch size~$1$ the peak footprint is $8.6$~GB and the pipeline
takes $5.8$~s per image; batched parsing at batch size~$16$ (bit-identical predictions, so purely
a throughput option) trades memory for speed, reaching $3.35$~s per image at a $16.6$~GB peak.
Table~\ref{tab:latency} breaks the batched cost down by stage: open-vocabulary detection and LLM
synthesis dominate ($1.69$~s and $1.35$~s), box-prompted SAM adds $0.29$~s, and the typed executor
costs only about $20$~ms, so program execution is negligible against the frozen perception modules.
Our reproduction of the diffusion-based RSVG-ZeroOV takes about $100$~s per image on the same GPU.
\input{tables/tab_latency}
\input{figures/fig_efficiency}
Fig.~\ref{fig:efficiency} plots the throughput--memory trade-off across parsing batch sizes: it
saturates once detection dominates, and the single-card operating range ends at batch size $24$
before batch size $32$ exceeds the $24$~GB budget. The co-resident point is for one-image online
serving: keeping the parser and segmenter both in memory avoids the per-request model reload that
the staged batch-size-one path pays, so single-image throughput rises slightly at a cost of
$+3$~GB.

\input{figures/fig_rrsisd_qual}
\input{figures/fig_quad}
\input{tables/tab_efficiency}

\subsection{Explicit Execution vs Implicit Matching}
\label{sec:exp:c2}
\input{tables/tab_selector}
Under the same frozen detector and SAM ViT-L segmenter configuration, we vary the selector
(RRSIS-D val, $n{=}1740$; Table~\ref{tab:c1geom}). All methods use this same perception
configuration, although their query formulation and candidate selection may differ, so these
results compare complete grounding pipelines under a fixed perception stack. Explicit program
execution improves mIoU by $+12.06$ over the stronger of the two implicit selectors (detector
confidence at 46.17, above GeoRSCLIP region--text similarity at 45.22). Of this, the continuous field
algebra alone accounts for $+11.24$ (field-only at 57.41) and the discrete operators add a
further $+0.82$. The advantage lies on the
spatial and compositional strata ($+18.77$ image-relative, $+7.47$ object-anchored, $+8.41$
compositional) and is near-zero on pure attribute ($+0.89$), where no spatial field exists and
the field-only selector reduces to the detector ordering. The ordering image $>$ object follows anchor
availability: the image frame is always present, whereas an object anchor must be detected and
bound to the correct instance. The gain concentrates on the samples for
which a spatial constraint can be executed, which is consistent with a benefit from explicit
execution given the shared perception. It also scales with same-class clutter, the defining RRSIS
difficulty: over that same stronger implicit selector the advantage is only $+1.7$~mIoU when a single
candidate is detected, rising to $+12.4$ at two or three candidates and
$+16.8$ at four to six. Explicit execution helps precisely where multiple same-class instances
compete. Two further selectors under the same configuration sharpen the comparison. The rule-based
spatial selector applies one image-frame heuristic to the same candidates: a single directional cue
takes the extreme box on that axis, a corner cue the diagonal extreme, ``centre''/``near'' the box
closest to the image centre, and any relational, nested, or attribute-only expression falls back to the
largest box. The LLM ranker uses the same Qwen3-4B and the same candidates, listed as normalised
integer coordinates, and returns one candidate index directly, so the only difference from the full
program is whether a typed program mediates the choice. They reach $53.91$ and $46.73$~mIoU, both
below the field-only special case at $57.41$ and the full program at $58.23$. The rule-based selector is
competitive where a single frame cue applies, reaching $62.08$ on image-relative against $63.34$ for
field-only, and falls off to $34.21$ on object-anchored and $32.39$ on compositional expressions, which
a fixed rule set cannot represent; neither hand-crafted spatial rules nor unconstrained LLM
ranking matches typed execution.

\subsection{Discrete Operators and Per-Stratum Gains}
\label{sec:exp:ops}
\input{tables/tab_risbench_ops}
The bottom rows of Table~\ref{tab:c1geom} compare the full
program with the field-only special case under the same configuration. On RRSIS-D val they add
$+0.82$ overall (paired $95\%$ CI $[+0.05,+1.58]$), concentrated on object-anchored ($+2.45$)
and compositional ($+3.42$) expressions and neutral on image-relative ($-0.10$), as expected
since a pure directional field already handles the frame-relative case. Table~\ref{tab:risbench_ops}
resolves the gain by stratum on RISBench, where the operators are exercised at scale. The gain is
$+1.13$ overall ($95\%$ CI $[+0.64,+1.63]$) and is carried by the superlative stratum
($+2.22$, $[+0.52,+3.84]$, $n{=}2424$); on compositional the operators match the field
approximation to within noise ($+0.63$, $[-0.58,+1.86]$), and the ordinal stratum
($n{=}52$) is too small to determine a sign. We therefore do not claim a uniform
per-stratum improvement: the measurable accuracy gain concentrates on superlatives, while the
operators' role on the remaining strata is one of coverage, since the field-only special
case cannot represent these constructions at all
(Section~\ref{sec:method:discrete}). The best-box-IoU analysis below reports how much of each
stratum's available IoU the operators capture. Aggregate gains are
modest because the field-only special case already resolves most frame-relative expressions, but
on the constructions that require an operator the gap is large: across all $16{,}159$
RISBench test expressions, $297$ tagged as superlative or compositional have a field-only IoU
below $0.1$ and a full-program IoU above $0.9$, for example \emph{``the rightmost tennis
court''} ($0.00\!\to\!1.00$ via \textsc{Argmax} on $x$) and \emph{``the bottom-most tennis court''}
($0.00\!\to\!0.99$ via \textsc{Argmax} on $y$). The qualitative traces in Figs.~\ref{fig:rrsisd_qual}
and~\ref{fig:risbench_qual} show one program-resolved case per stratum.

\subsection{Best-Box-IoU Headroom}
\label{sec:exp:oracle}
\input{tables/tab_oracle}
The best-box-IoU proxy (Table~\ref{tab:c2oracle}) segments the best-IoU candidate in the detected
pool. It reports how much IoU the proposed boxes already make available, a diagnostic for candidate
availability; it does not measure program correctness. The candidate
pool already contains a box above $0.5$ IoU for $86.6\%$ of RRSIS-D and $81.8\%$ of RISBench
expressions, so detection recall is high on both benchmarks. On RRSIS-D
GeoSelect captures $86.4\%$ of this best-box IoU, and the best-box value itself reaches $68.16$,
just below the strongest supervised specialist RSRefSeg2 at $69.17$ and above the others.
Read only as a candidate-availability diagnostic, this motivates testing stronger in-domain proposals or
finer tiling on RRSIS-D; it does not establish an error-free-ranking ceiling. The
object-anchored stratum is the exception at $74.1\%$ capture, a residual same-side selection loss of the
kind in Fig.~\ref{fig:failure}. On RISBench the best-box IoU is much higher at $75.44$ yet capture
drops to $73.3\%$: the detected pool usually contains a usable box, but a larger share of it goes
uncaptured than on RRSIS-D, and that uncaptured share concentrates on the new strata. Pure
attribute captures $88.6\%$, while superlative at $64.9\%$, compositional at $64.5\%$, and ordinal
at $45.9\%$ on $n{=}52$ leave the most room. Since a usable box is usually present on these strata, the
natural place to look for further training-free gains on RISBench is the program path: axis
estimation, anchor binding, and ranking on compositional expressions.
The proxy reports candidate availability and does not decompose the error causally, so we read this
as an investigation priority (Section~\ref{sec:discussion}).

\subsection{Reliability}
\label{sec:exp:reliability}
Per-expression synthesis is made safe by the reliability ladder of
Section~\ref{sec:method:reliability}. The program rate is the fraction of
expressions resolved by an executed program with a non-empty result, the rest falling back. It is
$83.4\%$ on RRSIS-D test, or $2904$ of $3481$, and $91.0\%$ on RISBench test, or $14698$ of $16159$.
The lower RRSIS-D figure comes from
legal programs whose candidate set is empty: the open-vocabulary detector returns no proposal for the
queried phrase. The remaining expressions fall back to
the field-only selector and recover its result exactly, and in aggregate the full system stays
above this field-only special case by $+0.42$ on RRSIS-D test and $+1.13$ on RISBench
(Table~\ref{tab:risbench_ops}). The program rate measures successful program resolution, that is,
synthesis, checking, execution, and a non-empty result, rather than correctness, and the two are
reported separately: a
0.6B parser reaches the highest program rate yet the lowest accuracy
(Section~\ref{sec:exp:parser}). A blind manual check of $100$ legal-program outputs on RRSIS-D val,
stratified across operator types to cover the rare strata, finds $98\%$ of the synthesised programs to
be correct interpretations, the two exceptions being size comparatives outside the spatial scope
(Section~\ref{sec:discussion}). The same check on $100$ RISBench val outputs finds $92\%$
correct. The eight errors are two size superlatives resolved by position alone, two dropped
positional cues, one wrong target noun, one wrong direction, one positional phrase used as an object
anchor, and one spurious nesting level. These two numbers are not
directly comparable to the program rate: the manual audit measures semantic correctness conditional on
a stratified sample of legal-program outputs, whereas the program rate measures how often synthesis,
checking, and execution complete with a non-empty result over the full split. In these stratified
samples of legal-program outputs, semantic correctness was $98\%$ on RRSIS-D val and $92\%$ on RISBench val.

\subsection{Parser Capacity and Architecture}
\label{sec:exp:parser}
\input{tables/tab_parser}
The synthesiser is a lightweight, swappable component (Table~\ref{tab:parser}). Within the
Qwen3 family, scaling is non-monotonic and peaks at 4B: the 0.6B parser loses $8.7$~mIoU despite
the highest program rate ($94.4\%$), and the 8B parser ($54.28$) sits below the 4B, so
the task rewards reliable schema following over raw capacity. Across architectures, a
dense transformer (Phi-4-mini, $57.05$) and a state-space model (Falcon-Mamba, $55.12$) both
stay within about $3$~mIoU of the Qwen3-4B headline, and Llama-3.2-3B reaches $54.90$ even with
a low $45.1\%$ program rate because the reliability ladder routes its many ill-formed programs to
the field-only fallback. The method therefore requires a parser of at least about $2$B that
follows the DSL schema reliably, but not a specific architecture: dense transformers and a
state-space model are interchangeable at the headline. A vision-language synthesiser does not raise
the headline either (bottom block of Table~\ref{tab:parser}): Qwen3-VL-4B~\cite{qwen3vl} scores
$56.73$~mIoU with and without the image, below the text-only Qwen3-4B.

\subsection{Mask Decoder}
\label{sec:exp:decoder}
The box-to-mask stage is replaceable. We use SAM ViT-L in single-mask mode, size-aligned
($308$~M) with the alternatives. Swapping only this stage on RRSIS-D val, with synthesis,
detection, and selection frozen and every decoder in single-mask mode, gives SAM~ViT-L
58.23, SAM~2.1 hiera-large 56.26, and SAM3 with a box prompt 53.22.
The decoder moves the headline within a $5.0$-point band, far smaller than the $+12.1$ that
explicit execution adds over implicit matching (Section~\ref{sec:exp:c2}), so the result is not
tied to a particular segmenter. SAM~ViT-L edges SAM~2.1 because on large referents SAM~2.1's
single mask tends to return a part instead of the whole object; SAM3, designed for short
concept prompts, does not surpass it. SAM~ViT-L is selected on val and used for the reported
test result.

\subsection{Parameter Analysis}
\label{sec:exp:param}
\input{tables/tab_param}
Centre scoring (Eq.~\ref{eq:boxscore}) collapses the tuning surface. Each candidate is
scored at a single point, making the selection invariant to any strictly increasing reshaping of a
field, so the directional exponent $\gamma$, the centre width $\sigma_c$, and the proximity
length $L_A$ change no prediction. Sweeping each over a wide range
($\gamma\!\in\![0.5,2.0]$, $\sigma_c\!\in\![0.15,0.35]$) leaves the headline at $58.23$~mIoU to
the reported precision, an exact confirmation of the analytic invariance of
Section~\ref{sec:method:fields} rather than an insensitivity that tuning could exploit. The
extremal temperature $\beta$ is likewise inert on this
split, which has essentially no superlative expressions for it to act on. The parameters
that remain (Table~\ref{tab:param}) are the scoring rule and the candidate set.

The scoring rule confirms the centre-sampling design: centre ($58.23$) edges region-mean
($58.07$) and region-max ($57.90$). The gain concentrates on the object-anchored and
compositional strata, where large referents make the region mean dilute the prior. Pure-attribute
is identical across all three rules because a uniform field reduces every rule to the detector
ordering. The detection threshold defines a flat plateau ($57.2$/$58.2$/$58.6$/$57.8$ at
$0.10$/$0.15$/$0.20$/$0.25$) on which the frozen $0.15$ sits within $0.4$~mIoU of the
val-optimal $0.20$; we keep the pre-registered value rather than re-select the val maximum, a
sub-CI difference that would not justify re-running the test and transfer suite. The SAHI tile
shows the only sharp dependence: $768$ ($58.23$) and $1024$ ($58.27$) are equivalent, but $512$
collapses to $46.76$ because tiles below the referent scale starve detection of recall,
confirming the frozen $768$ is well chosen and that the candidate set, not the field algebra, is
the sensitive component.

\subsection{Pretraining-Exposure Audit}
\label{sec:exp:c3}
\input{tables/tab_leakage}
\input{tables/tab_unseen_categories}
\input{tables/tab_risbench}
The open-vocabulary detector is pretrained on LAE-1M~\cite{laedino}, which overlaps RRSIS-D's
source imagery (both draw on DIOR, with LAE-1M including DIOR trainval while holding out its
test split). We quantify this exposure at the image-id level under two seen-set definitions
(Table~\ref{tab:c3leak}); the membership is detector-level and pipeline-independent, so the
audit re-buckets GeoSelect's own per-sample IoUs. On the exact gold seen-set the
seen$-$unseen gap is $+1.27$~mIoU and the unseen subset ($58.44$) is close to the full
result. We read this descriptively: on this image-id audit the difference is small, though it does not
rule out other forms of pretraining exposure. The conservative seen-set is a worst case (it
marks every RRSIS-D test image in DIOR trainval as seen regardless of which LAE-1M actually
sampled), and its unseen subset still reaches $54.46$. Both unseen subsets remain well above
prior training-free full-test results. An in-domain detector remains important for recall:
LAE-DINO reports that generic open-vocabulary detectors with natural-scene pretraining reach
only about $1$~AP$_{50}$ on DIOR, against $82$--$86$ once pretrained on LAE-1M. Within
GeoSelect the detector supplies category- and attribute-level proposals; the referring
decision is made by the executed program.

We audit RRSIS-D only. RISBench is built from DIOR and DOTA-v2~\cite{crobim}; the same DIOR and
DOTA imagery is among LAE-1M's largest training sources~\cite{laedino}, so its detector exposure
is plausibly at least as high as RRSIS-D's, and we make no exposure-robustness claim for RISBench. An image-id-level audit is
moreover not feasible on the released RISBench: its imagery is re-encoded as renamed
$512\times512$ crops without back-references to the source DIOR/DOTA-v2 ids, and the
correspondence between RISBench and source-dataset splits is undocumented, so neither id matching
nor content hashing against the originals is reliable. This exposure question is, however,
orthogonal to the attribution of our gain: the comparison of Section~\ref{sec:exp:c2}
uses the same frozen detector and segmenter for every selector, so its
$+12.06$~mIoU advantage is not explained by detector quality or by any
pretraining exposure, both shared across selectors. Pairing the executor with the strongest available
aerial-pretrained open-vocabulary detector is a deliberate part of the training-free design. The audit bears only on the absolute headline, where the
RRSIS-D gold-set gap of $+1.27$~mIoU is small on this measure.

\subsection{Cross-Dataset Transfer: RISBench}
\label{sec:risbench}
\input{figures/fig_risbench_qual}
\input{figures/fig_risbench_quad}
The full configuration is selected before test evaluation and applied to RISBench with no
test-set recalibration, so Table~\ref{tab:risbench} measures transfer. GeoSelect reaches 55.27~mIoU /
55.88~oIoU on RISBench test (mIoU $95\%$ bootstrap CI $[54.64,55.89]$), $1.74$ times the best
prior training-free method in mIoU and
$2.1$ times in oIoU, at about $88\%$ of RMSIN. With no RISBench supervision it surpasses the
early LSTM-based supervised methods RRN, BRINet, and CMPC+ on both metrics. The gain decomposes
into the in-domain candidate pipeline and a further $+1.13$ from the discrete operators, larger
than on RRSIS-D because RISBench is rich in the constructions they target
(Section~\ref{sec:exp:ops}). The two-dataset scope matches recent practice: RSVG-ZeroOV and the
concurrent ProVG~\cite{provg} evaluate on exactly these two.
Figures~\ref{fig:risbench_qual} and~\ref{fig:risbench_quad} show qualitative predictions across
strata and the five-way baseline comparison; the gap to the implicit baselines widens on
RISBench's longer, more constrained expressions.

\subsection{Failure Modes by Execution Route}
\label{sec:exp:failure}
\input{figures/fig_failure}
Every intermediate is inspectable, so we can trace each outcome to an execution path
(Fig.~\ref{fig:failure}). We partition every expression by the route that produced its final mask:
\emph{Success} covers IoU at least $0.5$; \emph{Detection} covers expressions where no candidate
reaches IoU~$0.5$; \emph{Selection} covers those where the field-only fallback is used yet the result
stays below $0.5$; and \emph{Program-path} covers those where an executed program is used yet the
result stays below $0.5$. Fig.~\ref{fig:failstats} reports the share of each outcome. \textbf{Detection} (a): the referent is too small to be proposed, so the
candidate set is empty and no prior, program, or mask can recover it ($\textsc{best-box}=0$);
this is an unrecoverable proposal-recall failure, and higher-resolution tiling or a stronger
detector directly targets this route. \textbf{Selection} (b): a near-perfect candidate exists (best-box IoU $0.85$) but
several instances satisfy ``on the left'' and the monotone ramp scores the furthest-left one
rather than the annotated instance. A graded directional field cannot encode whether ``on the
left'' denotes a region or an extremum (the monotone-versus-extremal distinction of
Section~\ref{sec:method:kernels}), so this ambiguity is a limitation of the field representation
rather than a tunable error.
\textbf{Program-path} (c,~d): the correct box is present (best-box IoU $0.94$/$0.66$) but the
executed program selects the wrong instance. In (c) the program is well-formed and keeps the
nested anchor for ``above the court on the lower right'', yet the resolution binds to the wrong
tennis court; in (d) a multi-cue phrase collapses to a single predicate. Such program-path errors,
which span synthesis, execution, and ranking, are the most frequent unsuccessful
outcome on RISBench (Fig.~\ref{fig:failstats}), where the detected pool usually already contains the
right box. The mask decoder contributes comparatively little: swapping it moves the headline within
a $5.0$-point band (Section~\ref{sec:exp:decoder}).

%% file: tables/tab_datasets.tex
\begin{table}[!t]
\caption{Statistics of the two evaluation benchmarks. Both are single-instance
referring-segmentation benchmarks with pixel masks; counts are image--expression--mask triplets,
one per referring expression.}
\label{tab:datasets}
\centering\footnotesize
\begin{tabular}{lcccccc}
\toprule
\multirow{2}{*}{Benchmark} & \multirow{2}{*}{Triplets} & \multirow{2}{*}{Categories} & \multirow{2}{*}{Image size} & \multicolumn{3}{c}{Split (train / val / test)} \\
\cmidrule(lr){5-7}
 & & & & Train & Val & Test \\
\midrule
RRSIS-D~\cite{rmsin}   & 17{,}402 & 20 & $800\!\times\!800$ & 12{,}181 & 1{,}740 & 3{,}481 \\
RISBench~\cite{crobim} & 52{,}472 & 26 & $512\!\times\!512$ & 26{,}300 & 10{,}013 & 16{,}159 \\
\bottomrule
\end{tabular}
\vspace{2pt}\footnotesize

RRSIS-D expressions carry 7 attribute types; RISBench carries 8 and averages 14.31 words per
expression (spatial resolution 0.1--30\,m). All design choices are frozen on RRSIS-D val; the
headline uses each benchmark's test split.
\end{table}

%% file: tables/tab_main.tex
\begin{table}[!t]
\caption{Main results on RRSIS-D test ($n{=}3481$). Fully-supervised specialists and trained
vision--language models are an upper reference, not training-free competitors; sources and
evaluation protocol in Section~\ref{sec:exp:setup}. \textbf{Bold}: best per method group; (--) metric not reported.}
\label{tab:main}
\centerline{%
\begin{tabular}{llcc}
\toprule
Method & Venue & mIoU & oIoU \\
\midrule
\multicolumn{4}{l}{\emph{Fully-supervised specialists}}\\
LAVT~\cite{lavt}              & CVPR'22  & 61.12 & 76.48 \\
LGCE~\cite{lgce}              & TGRS'24  & 60.98 & 76.33 \\
RMSIN~\cite{rmsin}            & CVPR'24  & 63.38 & 76.55 \\
FIANet~\cite{fianet}          & TGRS'24  & 64.01 & 76.81 \\
CroBIM~\cite{crobim}          & arXiv'25 & 64.24 & 76.37 \\
ProVG~\cite{provg}            & arXiv'26 & 65.44 & 77.62 \\
BTDNet~\cite{btdnet}          & arXiv'25 & 66.04 & 79.23 \\
RSRefSeg2~\cite{rsrefseg2}    & TGRS'26  & \textbf{69.17} & \textbf{79.45} \\
\midrule
\multicolumn{4}{l}{\emph{Trained vision--language models}}\\
GeoGround~\cite{geoground}    & arXiv'24 & 60.50 & --    \\
Text4Seg++~\cite{text4seg}    & TPAMI'26 & 62.80 & 74.20 \\
SegEarth-R1~\cite{segearthr1} & arXiv'25 & 66.40 & \textbf{78.01} \\
Sosa et al.\ (LoRA)~\cite{trainfreeseg} & arXiv'26 & \textbf{67.6} & -- \\
\midrule
\multicolumn{4}{l}{\emph{Training-free}}\\
SAM3 (sentence prompt)~\cite{sam3} & arXiv'26 & 15.50 & 16.87 \\
EKP-HRM~\cite{ekphrm}         & GRSL'26  & 19.21 & 20.68 \\
DGL-RSIS~\cite{li2026dgl}     & JAG'26   & 21.50 & --    \\
Sosa et al.\ (zero-shot)~\cite{trainfreeseg} & arXiv'26 & 24.9 & -- \\
RSVG-ZeroOV~\cite{rsvgzeroov} & AAAI'26  & 28.35 & 22.83 \\
\midrule
\textbf{GeoSelect (ours)}     & --       & \textbf{58.86} & \textbf{59.48} \\
\bottomrule
\end{tabular}}
\vspace{2pt}\footnotesize
The field-only special case of our method and the discrete-operator gain are reported separately
in the ablation of Table~\ref{tab:c1geom} (Section~\ref{sec:exp:c2}). SAM3 is a concept
segmenter; the full expression as prompt returns no instance for $69\%$ of cases (15.50), and
prompt formats are compared in Section~\ref{sec:exp:decoder}. Training-free numbers are quoted
from each source paper. The two Sosa et al.~\cite{trainfreeseg} rows are its fully training-free
(GPT-5 zero-shot) and LoRA-tuned (Qwen3-VL-2B) variants; only the former is training-free
(Section~\ref{sec:related}).
\end{table}

%% file: tables/tab_latency.tex
\begin{table}[!t]
\caption{Per-stage inference latency on a single NVIDIA GeForce RTX~3090 (RRSIS-D val, $n{=}200$,
no caching, parsing batch size 16). Program execution is negligible against
the frozen perception modules.}
\label{tab:latency}
\centering\footnotesize
\begin{tabular}{lc}
\toprule
Stage & Time per image \\
\midrule
LLM synthesis (Qwen3-4B)              & $1.35$~s \\
Open-vocabulary detection (LAE-DINO)  & $1.69$~s \\
Program execution (typed executor)    & $0.02$~s \\
Segmentation (SAM ViT-L)              & $0.29$~s \\
\midrule
Total                                 & $3.35$~s \\
\bottomrule
\end{tabular}
\end{table}

%% file: figures/fig_efficiency.tex
\begin{figure}[t]
\centering
\includegraphics[width=\columnwidth]{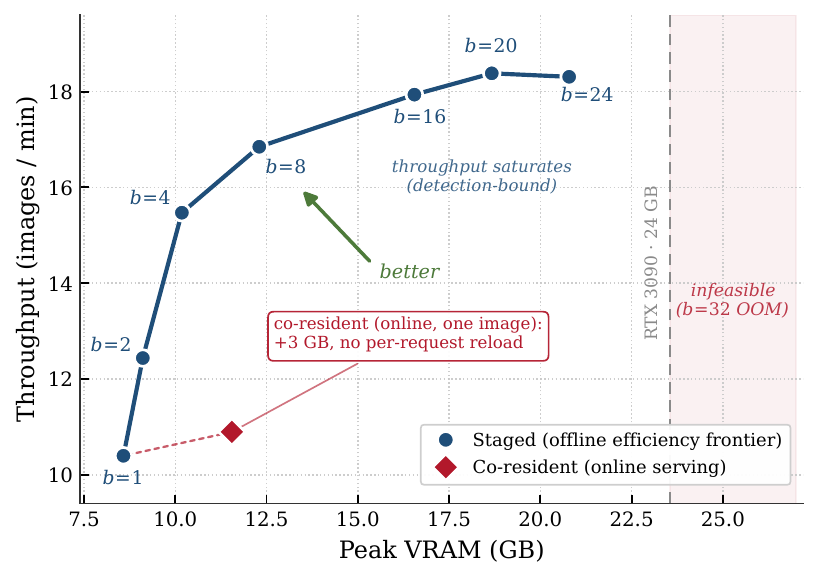}
\caption{Deployment efficiency on a single RTX~3090 (RRSIS-D val, $n{=}200$, no caching):
throughput versus peak VRAM across parsing batch sizes, with the co-resident point for one-image
online serving.}
\label{fig:efficiency}
\end{figure}

%% file: figures/fig_rrsisd_qual.tex
\begin{figure*}[!tp]
\centering
\includegraphics[width=\textwidth]{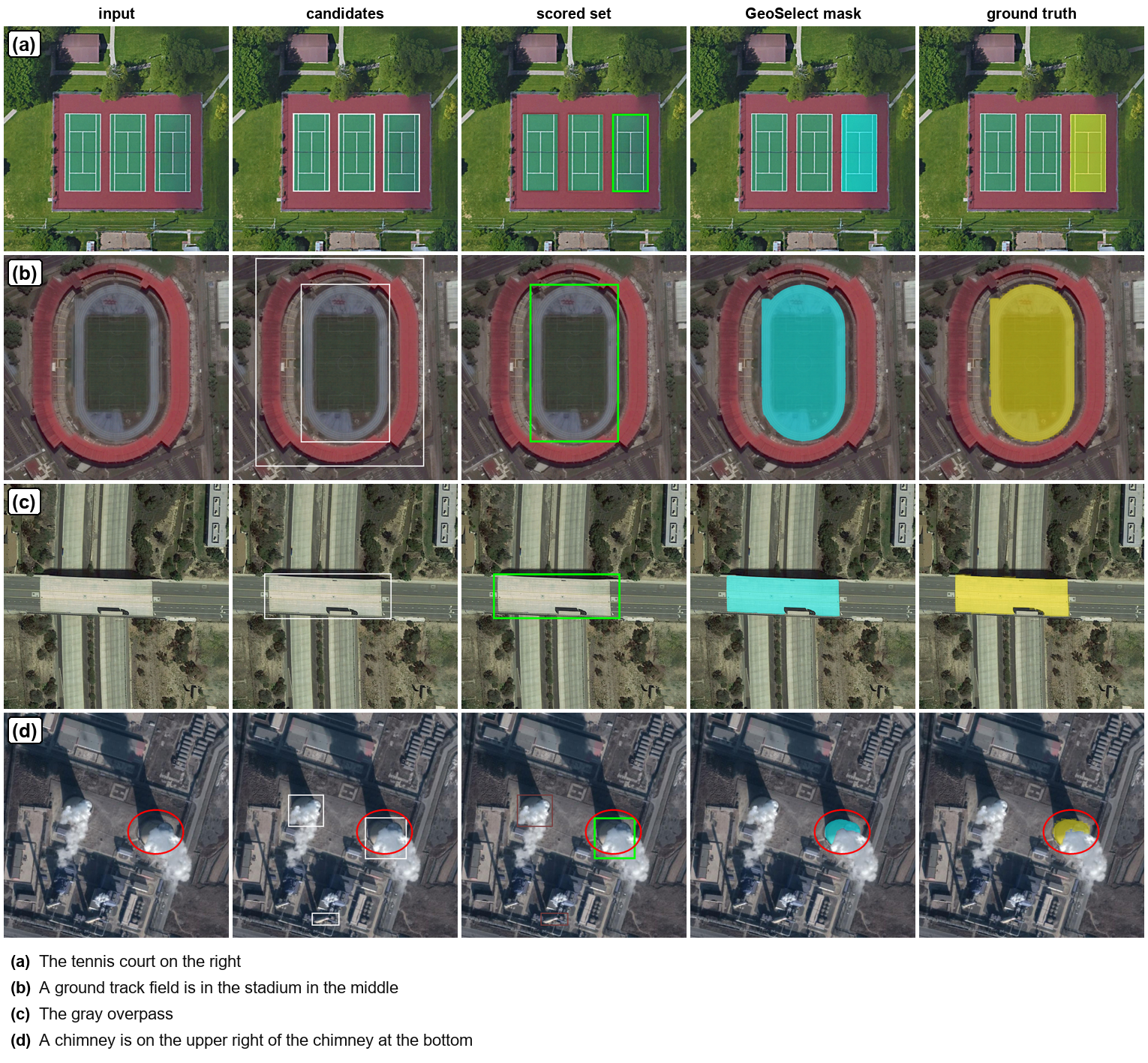}
\caption{\textbf{Pipeline traces on RRSIS-D test}, one case per row across the columns
input, candidate boxes, scored candidate set (brighter~$=$~higher score, chosen box in green),
GeoSelect mask (cyan), and ground truth (yellow). Rows span the strata: (a)~image-relative,
(b,~d)~object-anchored/compositional, (c)~attribute. Small targets are circled in red for
visibility. The referring expression of each row is printed below.}
\label{fig:rrsisd_qual}
\end{figure*}

%% file: figures/fig_quad.tex
\begin{figure*}[!tp]
\centering
\includegraphics[width=\textwidth]{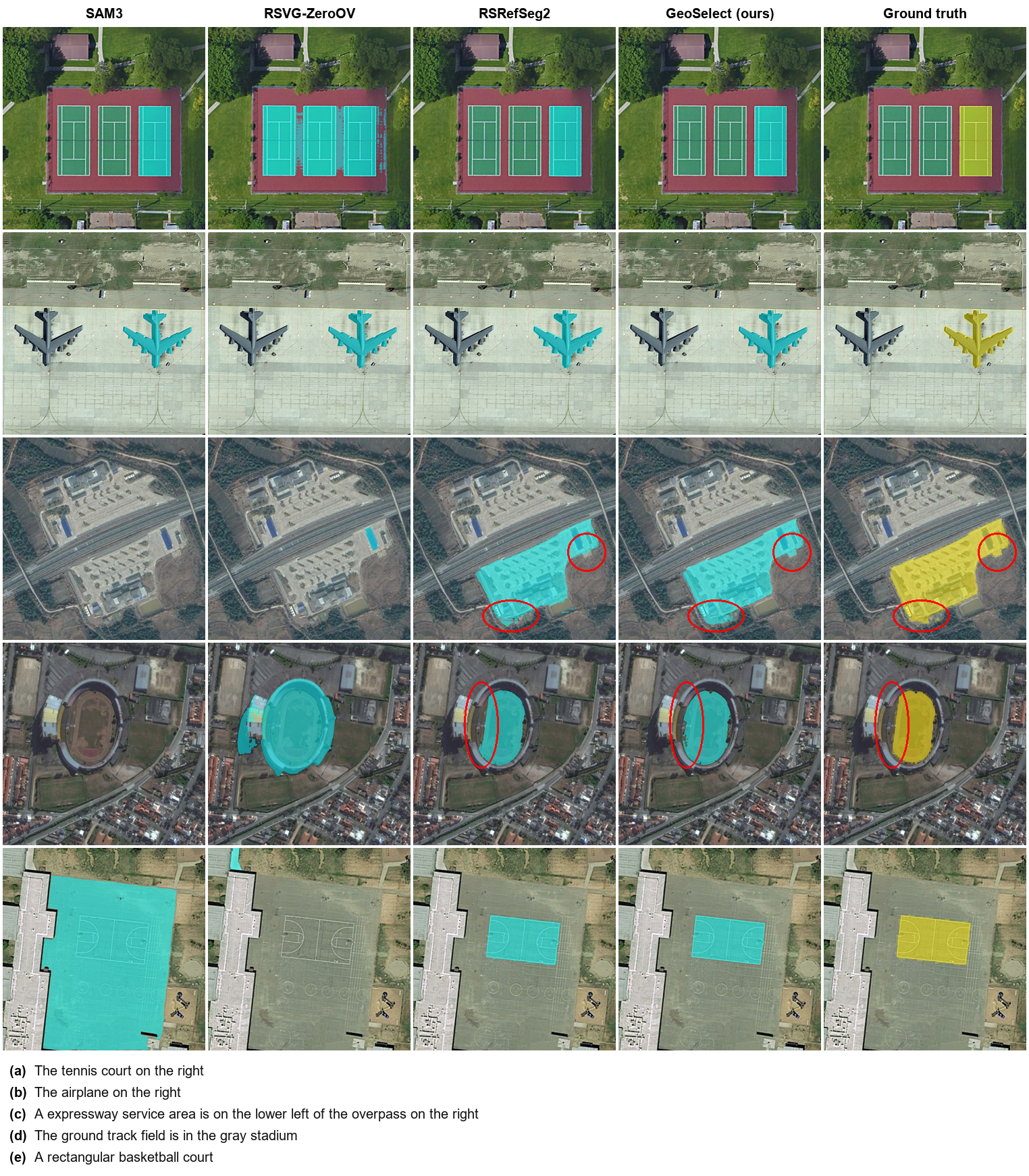}
\caption{\textbf{Qualitative comparison on RRSIS-D test.} Columns: SAM3~\cite{sam3}
(full-expression prompt), RSVG-ZeroOV~\cite{rsvgzeroov}, RSRefSeg2~\cite{rsrefseg2} (supervised),
GeoSelect (ours), ground truth. Predictions in cyan, GT in yellow. The training-free baselines
SAM3 and RSVG-ZeroOV over-segment multiple same-class regions or pick the wrong instance, whereas
GeoSelect, also training-free, resolves the queried instance comparably to the supervised
RSRefSeg2. Red ellipses mark the region that distinguishes the methods. The referring
expression of each row is printed below the panel.}
\label{fig:quad}
\end{figure*}

%% file: tables/tab_efficiency.tex
\begin{table}[!t]
\caption{Training cost. GeoSelect updates no weights and carries one frozen configuration across
datasets; the supervised methods fine-tune per dataset (RMSIN's $\sim$200\,M $=$ Swin-B $+$ BERT-base).}
\label{tab:efficiency}
\centerline{%
\begin{tabular}{llcc}
\toprule
Method & Paradigm & \shortstack{Trainable\\params} & \shortstack{Per-dataset\\training} \\
\midrule
RMSIN~\cite{rmsin}            & supervised        & full ($\sim$200\,M) & 40 epochs \\
RSRefSeg2~\cite{rsrefseg2}    & supervised, LoRA  & LoRA (PEFT)         & 300 epochs/dataset \\
RSVG-ZeroOV~\cite{rsvgzeroov} & training-free     & \textbf{0}          & none \\
\textbf{GeoSelect (ours)}     & training-free     & \textbf{0}          & none \\
\bottomrule
\end{tabular}}
\vspace{2pt}\footnotesize
RSRefSeg2 ships a separate checkpoint per benchmark; GeoSelect carries one. Inference latency is reported in the text.
\end{table}

%% file: tables/tab_selector.tex
\begin{table}[!t]
\caption{Selector comparison on RRSIS-D val ($n{=}1740$): all variants use the same frozen detector
and segmenter configuration, and differ in query formulation and selection mechanism.
IM/OBJ/ATTR/COMP: image-relative / object-anchored / pure-attribute / compositional. Detector
confidence and GeoRSCLIP are implicit; rule-based and LLM-ranking are simple explicit baselines; the
last two are ours, with and without the discrete operators.}
\label{tab:c1geom}
\centerline{%
\begin{tabular}{lccccc}
\toprule
Selector & mIoU & IM & OBJ & ATTR & COMP \\
\midrule
Detector confidence            & 46.17 & 44.47 & 33.80 & 55.76 & 32.75 \\
GeoRSCLIP~\cite{georsclip}     & 45.22 & 41.90 & 35.23 & 56.50 & 34.57 \\
Rule-based spatial             & 53.91 & 62.08 & 34.21 & 49.57 & 32.39 \\
LLM candidate ranking          & 46.73 & 46.26 & 35.33 & 53.61 & 33.60 \\
Ours, fields only              & 57.41 & \textbf{63.34} & 40.25 & 55.76 & 39.56 \\
\textbf{Ours, full program}    & \textbf{58.23} & 63.24 & \textbf{42.70} & \textbf{57.39} & \textbf{42.98} \\
\midrule
$\Delta$ vs.\ best implicit    & $\mathbf{+12.06}$ & $+18.77$ & $+7.47$ & $+0.89$ & $+8.41$ \\
$\Delta$ discrete vs.\ fields  & $+0.82$ & $-0.10$ & $+2.45$ & $+1.63$ & $+3.42$ \\
\bottomrule
\end{tabular}}
\vspace{2pt}\footnotesize
Explicit execution adds $+12.06$~mIoU over the best implicit selector (paired $95\%$ CI
$[+10.37,+13.77]$); the discrete operators add a further $+0.82$ (CI $[+0.05,+1.58]$). The
rule-based and LLM-ranking baselines are simple explicit references that trail both variants on every
stratum. Per-stratum operator effects on RISBench are in Table~\ref{tab:risbench_ops}.
\end{table}

%% file: tables/tab_risbench_ops.tex
\begin{table}[!t]
\caption{Discrete-operator gain by stratum on RISBench test ($n{=}16159$): the field-only
special case against the full program on the \emph{same} candidates. $\Delta$ is the paired mean
IoU gain with a $95\%$ bootstrap CI (Section~\ref{sec:exp:ops}).}
\label{tab:risbench_ops}
\centering
\begin{tabular}{lrccl}
\toprule
Stratum & $n$ & fields & full & $\Delta$ (95\% CI) \\
\midrule
Superlative   & 2424  & 49.28 & 51.50 & $+2.22$ [$+0.52,+3.84$] \\
Compositional & 3998  & 49.17 & 49.81 & $+0.63$ [$-0.58,+1.86$] \\
Ordinal       & 52    & 37.59 & 33.04 & $-4.55$ [$-17.3,\;+8.7$] \\
\midrule
Overall       & 16159 & 54.13 & 55.27 & $+1.13$ [$+0.64,+1.63$] \\
\bottomrule
\end{tabular}
\vspace{2pt}\par\footnotesize
$\Delta$ and its CI are computed from unrounded per-sample IoUs; the \emph{fields} and
\emph{full} columns are rounded independently, so their printed difference can differ from
$\Delta$ by $0.01$.
\end{table}

%% file: tables/tab_oracle.tex
\begin{table}[!t]
\caption{Per-sample best-box-IoU proxy: IoU from segmenting the best-IoU candidate in the
detected pool. \emph{actual} is GeoSelect; \emph{best} the proxy; \emph{cap.} the captured
fraction. Strata absent from a benchmark are omitted.}
\label{tab:c2oracle}
\centerline{%
\begin{tabular}{lcccc}
\toprule
Stratum & $n$ & actual & best & cap.\% \\
\midrule
\multicolumn{5}{l}{\emph{RRSIS-D test} ($n{=}3481$)}\\
overall       & 3481 & 58.86 & 68.16 & 86.4 \\
image         & 1853 & 64.05 & 70.08 & 91.4 \\
object        &  628 & 43.18 & 58.29 & 74.1 \\
attribute     & 1000 & 59.09 & 70.81 & 83.5 \\
compositional &  374 & 45.76 & 56.43 & 81.1 \\
\midrule
\multicolumn{5}{l}{\emph{RISBench test} ($n{=}16159$)}\\
overall       & 16159 & 55.27 & 75.44 & 73.3 \\
image         & 11488 & 54.59 & 75.54 & 72.3 \\
object        &  3949 & 55.62 & 75.70 & 73.5 \\
attribute     &   722 & 64.11 & 72.34 & 88.6 \\
superlative   &  2424 & 51.50 & 79.30 & 64.9 \\
ordinal       &    52 & 33.04 & 71.99 & 45.9 \\
compositional &  3998 & 49.81 & 77.23 & 64.5 \\
\bottomrule
\end{tabular}}
\vspace{2pt}\footnotesize
RRSIS-D test contains a negligible number of superlative/ordinal expressions
($n{\le}1$), so those strata are reported only on RISBench. Capture is computed from unrounded
mIoU. The ordinal row ($n{=}52$) is small and reads with wide intervals
(cf.\ the per-stratum CIs in Table~\ref{tab:risbench_ops}); it is reported with its sample
size and not over-interpreted.
\end{table}

%% file: tables/tab_parser.tex
\begin{table}[!t]
\caption{Synthesiser scaling, cross-architecture, and visual input on RRSIS-D val (frozen
configuration; swap only the synthesiser). \emph{prog\%} is the program rate: the fraction of
expressions resolved by an executed program with a non-empty result.}
\label{tab:parser}
\centering
\begin{tabular}{llccc}
\toprule
Synthesiser & Arch & mIoU & oIoU & prog\% \\
\midrule
\multicolumn{5}{l}{\emph{Qwen3 family scaling (non-monotonic; 4B is the sweet spot)}}\\
Qwen3-0.6B & dense Tf & 49.57 & 51.72 & 94.4 \\
Qwen3-1.7B & dense Tf & 56.30 & 58.09 & 82.1 \\
\textbf{Qwen3-4B} & dense Tf & \textbf{58.23} & \textbf{59.30} & 82.4 \\
Qwen3-8B & dense Tf & 54.28 & 54.70 & 78.4 \\
\midrule
\multicolumn{5}{l}{\emph{Cross-architecture ($\sim$3--7B)}}\\
\textbf{Qwen3-4B} & dense Tf & \textbf{58.23} & \textbf{59.30} & 82.4 \\
Phi-4-mini~\cite{phi4} & dense Tf & 57.05 & 57.93 & 78.8 \\
Falcon-Mamba-7B~\cite{falconmamba} & SSM & 55.12 & 57.18 & 81.7 \\
Llama-3.2-3B~\cite{llama32} & dense Tf & 54.90 & 55.84 & 45.1 \\
\midrule
\multicolumn{5}{l}{\emph{Visual input (text-only vs.\ vision-language, 4B class)}}\\
\textbf{Qwen3-4B} & dense Tf & \textbf{58.23} & \textbf{59.30} & 82.4 \\
Qwen3-VL-4B~\cite{qwen3vl} \small(text) & VLM & 56.73 & 57.38 & 80.0 \\
Qwen3-VL-4B \small($+$image) & VLM & 56.73 & 58.04 & 81.1 \\
\bottomrule
\end{tabular}
\vspace{2pt}\par\footnotesize
Giving Qwen3-VL-4B the image leaves mIoU unchanged ($56.73$), moves oIoU by $+0.66$, and moves the
program rate by $+1.1$ points. The text-only Qwen3-4B stays ahead of both vision-language variants
on mIoU.
\end{table}

%% file: tables/tab_param.tex
\begin{table}[!t]
\caption{Parameter analysis on RRSIS-D val ($n{=}1740$), one parameter varied around the frozen
value (\textbf{bold}). Field-shape parameters ($\gamma,\sigma_c,L_A$) are omitted: centre
scoring (Eq.~\ref{eq:boxscore}) makes selection invariant to them, so every value yields
$58.23$ exactly.}
\label{tab:param}
\centering
\begin{tabular}{llcc}
\toprule
Parameter & value & mIoU & oIoU \\
\midrule
\multicolumn{4}{l}{\emph{Scoring rule}}\\
\texttt{score\_method} & region-max  & 57.90 & 59.22 \\
                       & region-mean & 58.07 & 58.73 \\
                       & \textbf{centre} & \textbf{58.23} & \textbf{59.30} \\
\midrule
\multicolumn{4}{l}{\emph{Detection threshold}}\\
$\tau_{\text{det}}$ & 0.10 & 57.20 & 57.90 \\
                    & \textbf{0.15} & \textbf{58.23} & \textbf{59.30} \\
                    & 0.20 & 58.60 & 59.90 \\
                    & 0.25 & 57.80 & 59.84 \\
\midrule
\multicolumn{4}{l}{\emph{SAHI tile size}}\\
tile & 512  & 46.76 & 41.52 \\
     & \textbf{768}  & \textbf{58.23} & \textbf{59.30} \\
     & 1024 & 58.27 & 59.83 \\
\bottomrule
\end{tabular}
\end{table}

%% file: tables/tab_leakage.tex
\begin{table}[!t]
\caption{Image-id-level seen/unseen audit on RRSIS-D test (detector--benchmark overlap;
membership is detector-level and pipeline-independent, so the seen-sets apply unchanged and the
metrics reported here are GeoSelect's own). UNSEEN denotes samples outside the specified seen-set; the unseen
mIoU is the exposure-audited number.}
\label{tab:c3leak}
\centerline{%
\begin{tabular}{lccccc}
\toprule
\multirow{2}{*}{Seen-set} & \multirow{2}{*}{seen\%} & \multicolumn{2}{c}{SEEN} & \multicolumn{2}{c}{UNSEEN} \\
\cmidrule(lr){3-4}\cmidrule(lr){5-6}
 & & mIoU & oIoU & mIoU & oIoU \\
\midrule
Exact (gold) & 32.8 & 59.71 & 58.60 & \textbf{58.44} & 59.91 \\
Conservative & 54.3 & 62.56 & 61.47 & \textbf{54.46} & 56.35 \\
\bottomrule
\end{tabular}}
\vspace{2pt}\footnotesize
Two seen-set definitions: the exact gold set (images the detector actually trained on) and a
conservative worst case (every RRSIS-D test image in DIOR trainval marked seen). Analysis in
Section~\ref{sec:exp:c3}.
\end{table}

%% file: tables/tab_unseen_categories.tex
\begin{table}[!t]
\caption{Per-category mIoU on RRSIS-D test by pretraining exposure ($3{,}418$ category-mapped
expressions, $63$ program-null excluded; seen/unseen split from Table~\ref{tab:c3leak}).
\textbf{Bold}: the $9$ of $20$ categories where unseen meets or exceeds seen. The weighted unseen
mIoU here is $58.64$; the all-expression $58.44$ is in Table~\ref{tab:c3leak}.}
\label{tab:unseen_categories}
\centering\footnotesize
\begin{tabular}{lcccc}
\toprule
\multirow{2}{*}{Category} & \multicolumn{2}{c}{UNSEEN} & \multicolumn{2}{c}{SEEN} \\
\cmidrule(lr){2-3}\cmidrule(lr){4-5}
 & mIoU & $n$ & mIoU & $n$ \\
\midrule
Airplane & \textbf{75.97} & 54 & 66.90 & 45 \\
Airport & 53.65 & 160 & 57.57 & 36 \\
Baseball field & 63.72 & 135 & 75.30 & 94 \\
Basketball court & \textbf{73.06} & 76 & 71.05 & 44 \\
Bridge & \textbf{50.75} & 195 & 43.50 & 67 \\
Chimney & 74.22 & 75 & 77.58 & 32 \\
Dam & \textbf{61.18} & 133 & 53.75 & 42 \\
Expressway service area & 69.29 & 67 & 76.00 & 51 \\
Expressway toll station & 64.93 & 93 & 72.73 & 21 \\
Golf field & 54.37 & 87 & 55.96 & 24 \\
Ground track field & 53.61 & 168 & 55.82 & 66 \\
Harbor & 25.70 & 16 & 36.54 & 11 \\
Overpass & \textbf{55.18} & 158 & 48.38 & 63 \\
Ship & 66.73 & 115 & 69.68 & 62 \\
Stadium & \textbf{75.93} & 63 & 74.19 & 25 \\
Storage tank & \textbf{71.47} & 68 & 64.04 & 39 \\
Tennis court & 66.78 & 45 & 77.27 & 57 \\
Train station & \textbf{64.16} & 105 & 61.81 & 23 \\
Vehicle & 45.10 & 306 & 47.83 & 238 \\
Windmill & \textbf{55.44} & 185 & 55.34 & 74 \\
\bottomrule
\end{tabular}
\end{table}

%% file: tables/tab_risbench.tex
\begin{table}[!tb]
\caption{Cross-dataset transfer to RISBench test ($n{=}16159$) under the \emph{frozen} RRSIS-D
configuration, with no test-set recalibration. Supervised rows: RISBench leaderboard compiled in
RSRefSeg2~\cite{rsrefseg2}/CroBIM~\cite{crobim}. \textbf{Bold}: best per method group.}
\label{tab:risbench}
\centerline{%
\begin{tabular}{llcc}
\toprule
Method & Venue & mIoU & oIoU \\
\midrule
\multicolumn{4}{l}{\emph{Fully-supervised specialists} (RISBench leaderboard~\cite{crobim})}\\
RRN~\cite{rrn}                & CVPR'18  & 43.18 & 49.67 \\
BRINet~\cite{brinet}          & CVPR'20  & 42.91 & 48.73 \\
CMPC+~\cite{cmpcplus}         & TPAMI'21 & 46.73 & 53.98 \\
CRIS~\cite{cris}              & CVPR'22  & 55.18 & 69.11 \\
LAVT~\cite{lavt}              & CVPR'22  & 61.93 & 74.15 \\
RMSIN~\cite{rmsin}            & CVPR'24  & 63.07 & 74.09 \\
CARIS~\cite{caris}            & MM'23    & 65.79 & \textbf{75.10} \\
CroBIM~\cite{crobim}          & arXiv'25 & 67.32 & 73.61 \\
RSRefSeg2~\cite{rsrefseg2}    & TGRS'26  & \textbf{72.57} & 74.77 \\
\midrule
\multicolumn{4}{l}{\emph{Training-free}}\\
SAM3 (sentence prompt)~\cite{sam3} & arXiv'26 & 21.09 & 26.20 \\
RSVG-ZeroOV~\cite{rsvgzeroov} & AAAI'26  & 31.84 & 26.35 \\
\midrule
\textbf{GeoSelect (ours)}     & --       & \textbf{55.27} & \textbf{55.88} \\
\bottomrule
\end{tabular}}
\vspace{2pt}\footnotesize
The field-only special case of our method and the per-stratum discrete-operator gain
($+1.13$~mIoU overall) are reported in the ablation of Table~\ref{tab:risbench_ops}
(Section~\ref{sec:exp:ops}). SAM3 returns no instance for $64\%$ of cases here.
\end{table}

%% file: figures/fig_risbench_qual.tex
\begin{figure*}[!tp]
\centering
\includegraphics[width=\textwidth]{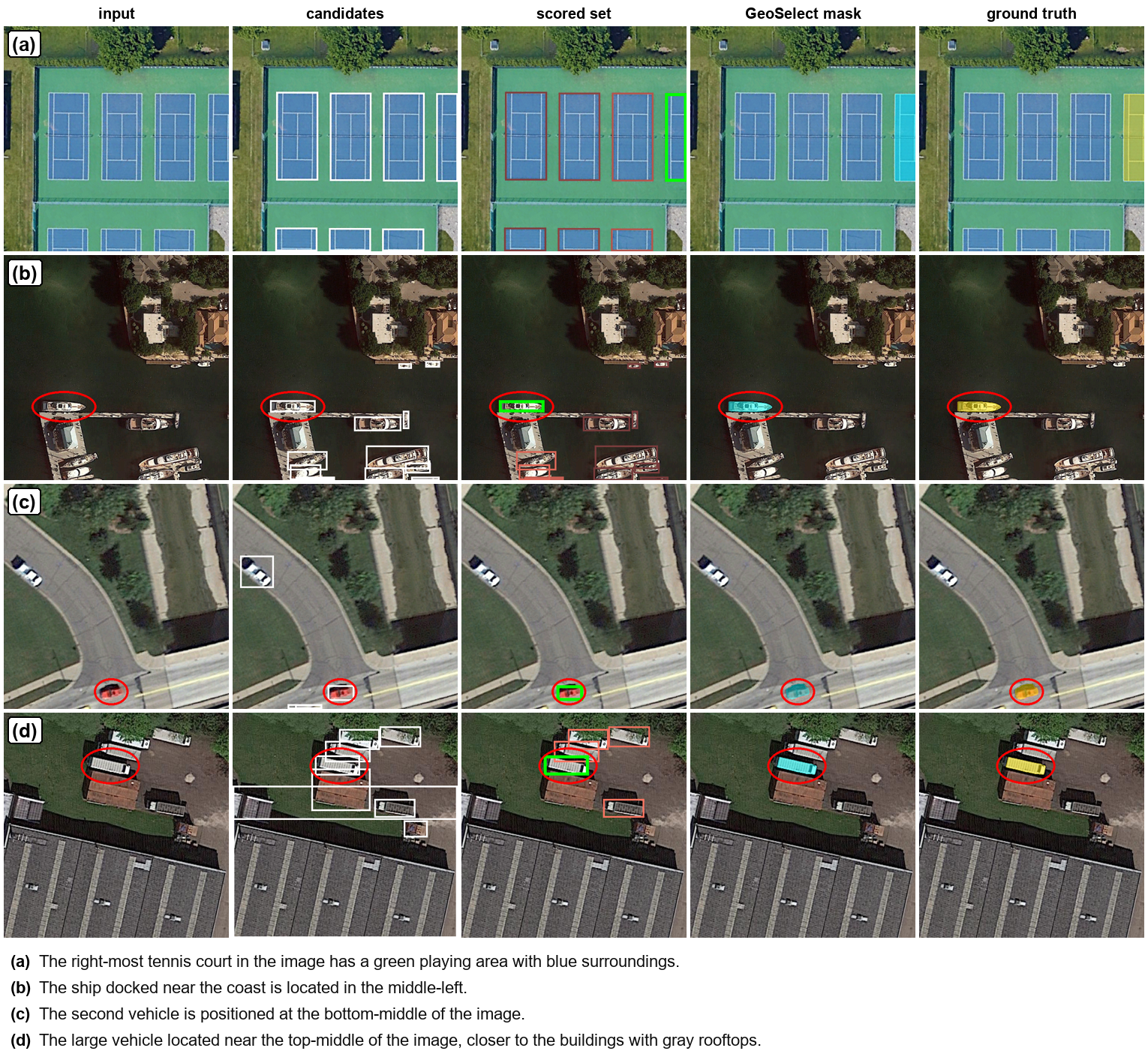}
\caption{\textbf{Pipeline traces on RISBench test} (frozen RRSIS-D configuration), one case per
row across input, candidate boxes, scored candidate set (brighter~$=$~higher score, chosen box in
green), GeoSelect mask (cyan), and ground truth (yellow). Rows span (a)~superlative,
(b)~compositional, (c)~ordinal (\emph{``the second vehicle''}), and (d)~object-anchored
proximity. Small targets are circled in red for visibility. The
referring expression of each row is printed below.}
\label{fig:risbench_qual}
\end{figure*}

%% file: figures/fig_risbench_quad.tex
\begin{figure*}[!tp]
\centering
\includegraphics[width=\textwidth]{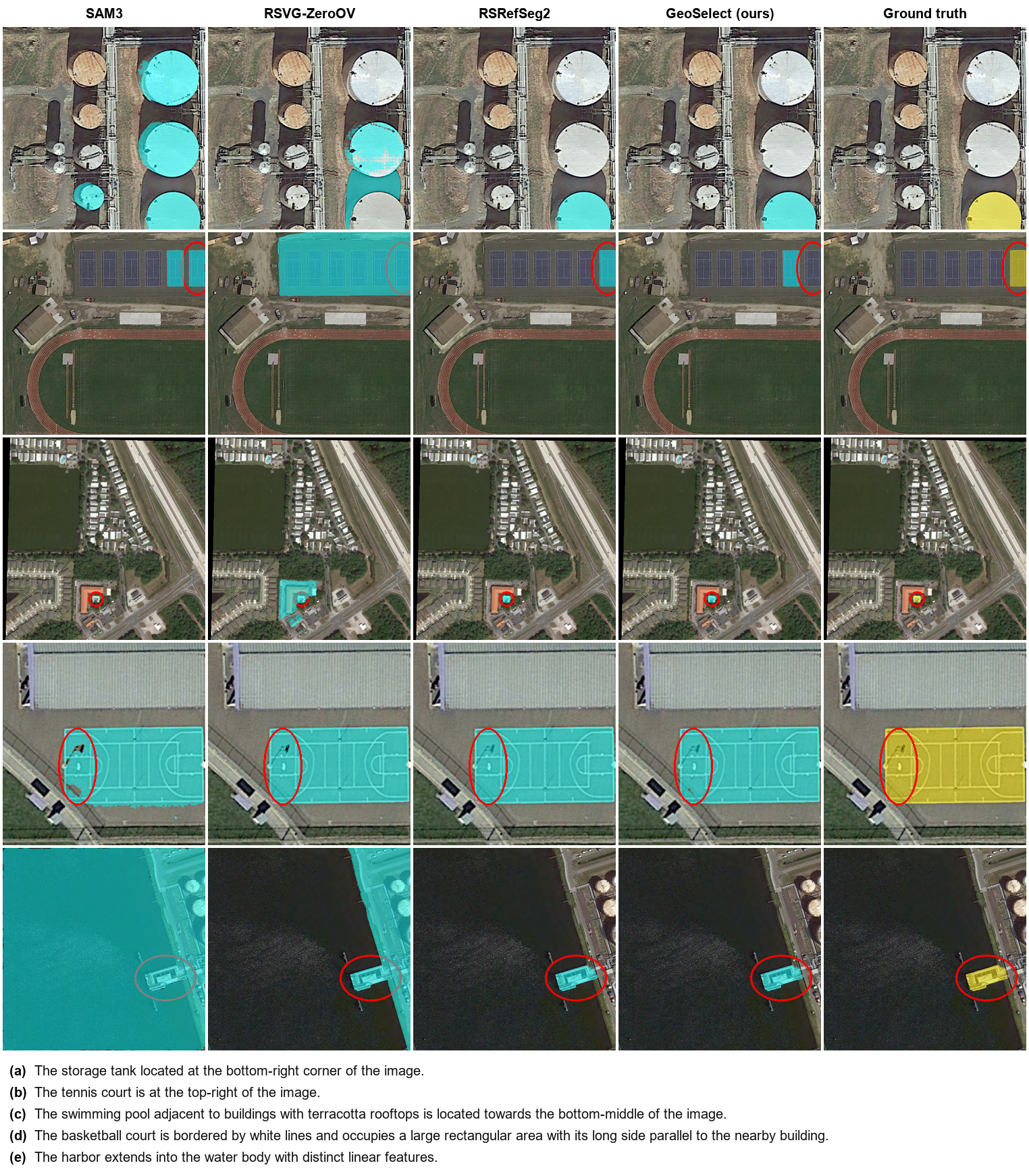}
\caption{\textbf{Qualitative comparison on RISBench test.} Columns: SAM3~\cite{sam3}
(full-expression prompt), RSVG-ZeroOV~\cite{rsvgzeroov}, RSRefSeg2~\cite{rsrefseg2} (supervised),
GeoSelect (ours), ground truth. Predictions in cyan, GT in yellow. The longer, more constrained RISBench
expressions widen the gap to the implicit baselines. Red ellipses mark a small or
method-distinguishing target region. The referring expression of each row is
printed below the panel.}
\label{fig:risbench_quad}
\end{figure*}

%% file: figures/fig_failure.tex
\begin{figure*}[!tp]
  \centering
  \includegraphics[width=0.86\textwidth]{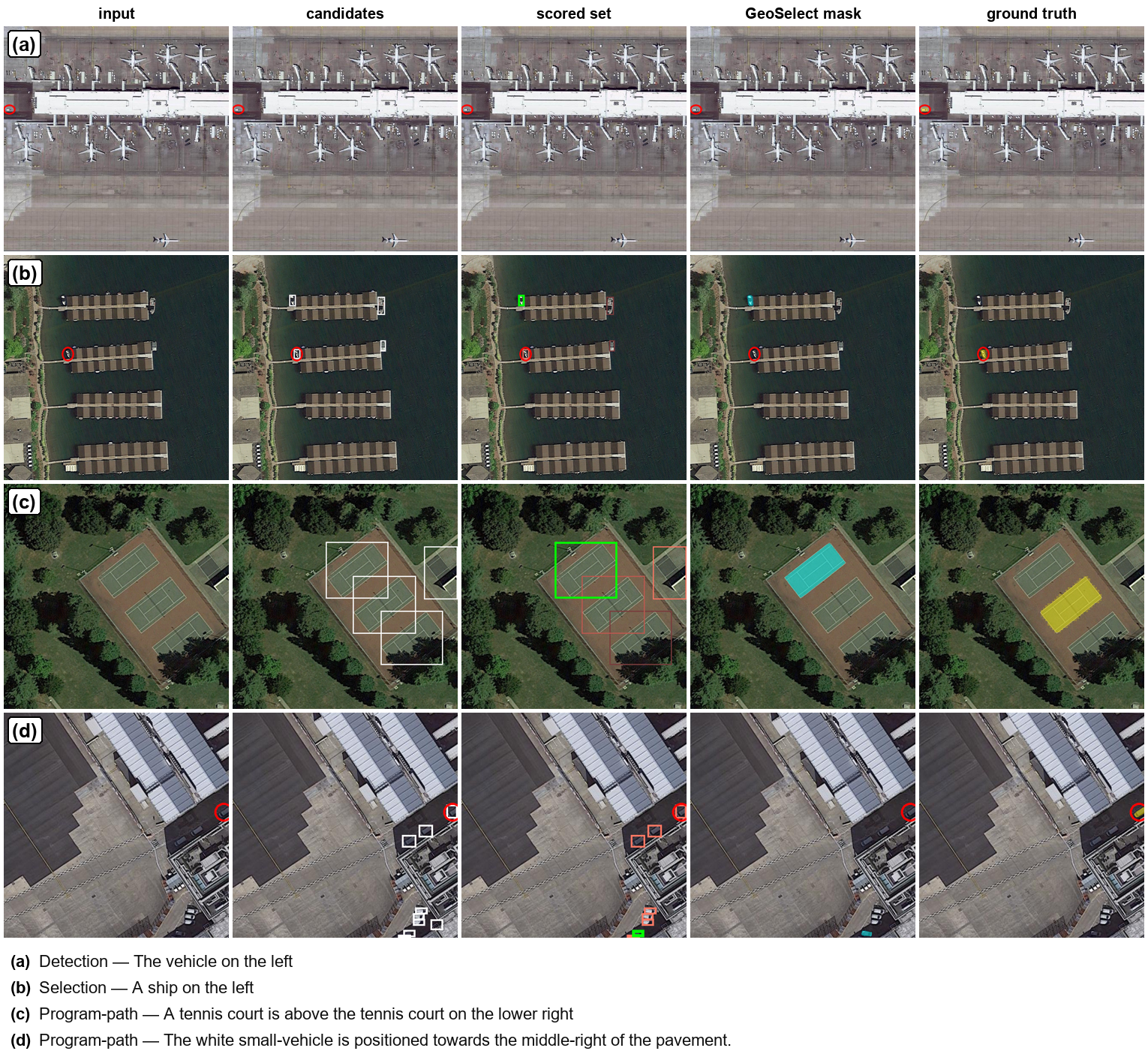}
  \caption{\textbf{Representative failure traces by execution route} (cyan~$=$~prediction, yellow~$=$~ground
  truth; small targets circled in red). \textbf{(a)~Detection}: the referent is too small to be
  proposed. \textbf{(b)~Selection}: a near-perfect candidate exists but the monotone ramp picks the
  wrong instance. \textbf{(c,~d)~Program-path}: the correct box is present but the executed program
  selects the wrong instance. Per-case expressions and IoUs are in Section~\ref{sec:exp:failure}.}
  \label{fig:failure}
\end{figure*}

%% file: sections/discussion.tex
\section{Discussion and Limitations}
\label{sec:discussion}

\input{figures/fig_failure_stats}
\paragraph*{Where the headroom is}
The best-box-IoU proxy (Section~\ref{sec:exp:oracle}) segments the best-IoU candidate in the detected
pool. It is a candidate-availability diagnostic: it reports how much IoU the proposals already make
available, and it is neither a per-sample upper bound nor a decomposition into synthesis, execution,
ranking, and mask decoding. Read as a directional signal, it gives a benchmark-specific picture of
where the remaining error sits. On
RRSIS-D the proxy leaves a $9.30$-point gap between the actual mIoU of $58.86$ and the best-box value
of $68.16$, and that best-box value itself lies just below the strongest supervised specialist; the
outcome partition in Fig.~\ref{fig:failstats} assigns $11.5\%$ of RRSIS-D expressions to detection and
$15.9\%$ to the program path. Proposal recall and program-path behaviour are therefore complementary
targets on RRSIS-D: a missed proposal is unrecoverable, while the program path accounts for the
larger share of unsuccessful outcomes.
On RISBench the balance shifts toward the program
path. Detection transfers well (the best-box IoU rises to $75.4$), but the proxy gap widens to $20.17$
points, the program-path share rises to $23.8\%$, and the uncaptured share concentrates on the
superlative, ordinal, and compositional strata, where the correct box is usually already among the
candidates. This makes the program path the clearer priority on RISBench (Section~\ref{sec:exp:oracle}).
The discrete operators move these strata well above the field-only floor, and the
$25$--$40$-point gap to the best-box IoU on them marks the most promising direction for further training-free
gains: better axis estimation under detector noise, anchor binding for nested expressions, and a
quantitative spatial-consistency check on the synthesised program. Because the proxy is non-causal, we
read these as investigation priorities rather than isolated error sources.

\paragraph*{Is the program machinery necessary?}
The field-only special case already doubles the best prior training-free result, which
invites the question of whether the DSL and its discrete operators earn their place. The answer
is yes, at two levels. The accuracy the field-only case reaches is itself a product of the same
executed representation. Under the same frozen detector and segmenter, executing a geometric program
beats the stronger of the two implicit selectors by $+12.06$~mIoU ($95\%$ CI $[+10.37,+13.77]$,
Section~\ref{sec:exp:c2}), consistent with explicit execution over the field algebra contributing the
bulk of the gain. On top of that, the discrete operators extend the same calculus to constructions the
field algebra structurally cannot express, and deliver a resolved $+2.22$~mIoU on RISBench
superlatives (Table~\ref{tab:risbench_ops}). The field-only number is thus not evidence against
the apparatus but a lower rung of it.

\paragraph*{The expressiveness--reliability trade-off}
Replacing a closed predicate lexicon with per-expression program synthesis buys the
compositional coverage of Section~\ref{sec:exp:ops} at the cost of programs that can be
ill-formed or semantically wrong. The well-formedness checker and the reliability ladder bound this
downside (Section~\ref{sec:exp:reliability}). This is the structural difference from prompt-era
visual-programming pipelines
that dispatch to opaque neural subroutines: our symbolic half is an explicit geometric
calculus whose every intermediate is inspectable and whose failures are catchable.

\paragraph*{Methodological scope}
The few-shot schema was developed on RRSIS-D val and includes grammar rules (extremal axis
polarity, image-quadrant anchoring, ``middle'' mapped to \textsc{Center}) that are general
language-to-program mappings rather than dataset-specific tuning, and the configuration
transfers to RISBench with no test-set recalibration. We report it as designed on RRSIS-D val and
validated on RISBench rather than as design-free, since some compositional constructions are
rare on RRSIS-D.

\paragraph*{Limitations}
GeoSelect targets the single-instance regime, exposed through the parser's cardinality field;
generalised $0/1/k$ referring~\cite{marnoto2025generalized} is out of scope. Absolute metric
thresholds, such as \emph{``within 50~m''}, require a known ground-sample distance and are also out
of scope. Size and
comparative-size expressions such as \emph{``smaller than''} are outside the claimed coverage: size
is orthogonal to the directional field algebra, and the only size signal available downstream of the
text-only synthesiser is the detected bounding-box area, an unreliable proxy for semantic size. A
non-comparative size adjective may remain in the detector-query noun phrase, but the system does not
rank instances by size.
It assumes an in-domain open-vocabulary
detector, without which proposal recall collapses (Section~\ref{sec:exp:c3}). The axis-aligned box
geometry underlying the extremal axes is orientation-sensitive for elongated objects
such as ships and bridges: on RISBench superlatives, IoU falls from $57.9$ for near-square
selected boxes (aspect ratio $\le\!1.5$) to $23.9$ for elongated ones (ratio $>\!3$, $n{=}298$).
Soft ranking mitigates but does not remove this; an oriented-box geometry is a clear next
step.
Finally, the ordinal stratum is small on current benchmarks ($n{=}52$ on RISBench), so its
large best-box-IoU gap, while suggestive, rests on too few samples to interpret.

%% file: figures/fig_failure_stats.tex
\begin{figure}[!t]
\centering
\includegraphics[width=\columnwidth]{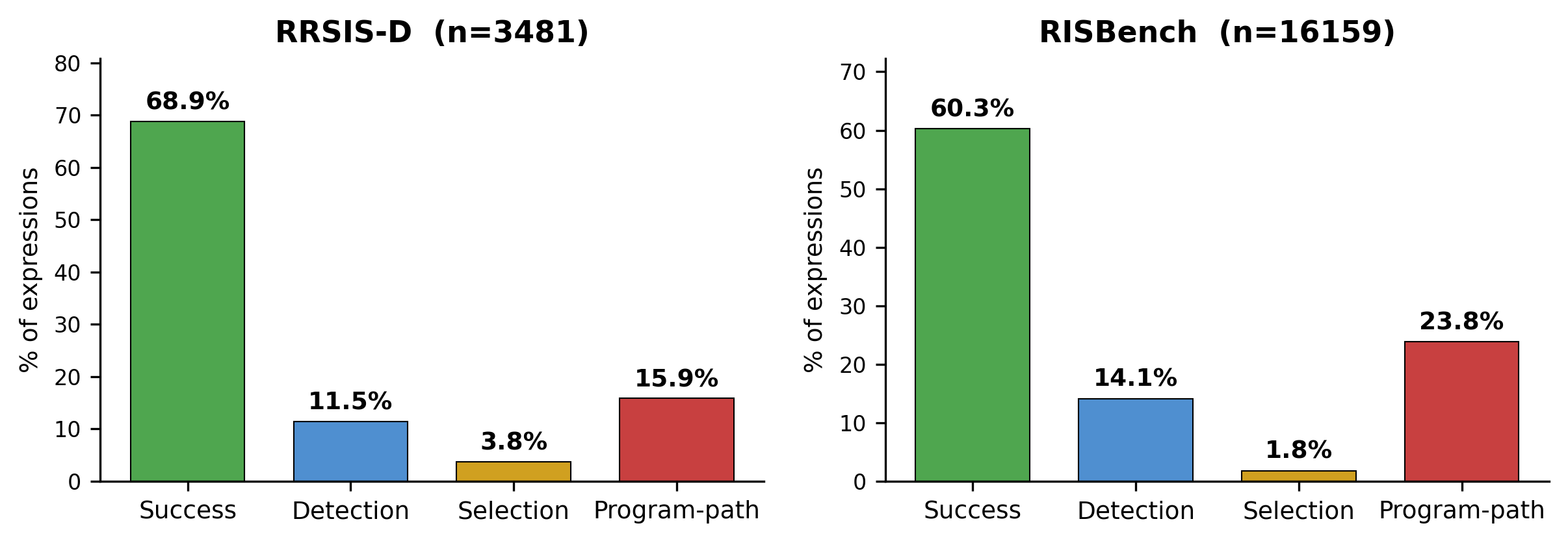}
\caption{\textbf{Outcome partition by execution route} on the two test sets, using the
\emph{Success}/\emph{Detection}/\emph{Selection}/\emph{Program-path} outcomes defined in
Section~\ref{sec:exp:failure}. Each names the route that produced the reported mask, not a
causal stage diagnosis; their cross-dataset shift is read in Section~\ref{sec:discussion}.}
\label{fig:failstats}
\end{figure}

%% file: sections/conclusion.tex
\section{Conclusion}
\label{sec:conclusion}
We reframed training-free referring remote sensing image segmentation as explicit
spatial-program execution. A frozen, text-only language model synthesises a typed program in a
small DSL, a well-formedness checker accepts it, and a deterministic executor runs it over a single
scored-candidate-set type that unifies continuous geometric fields with discrete set and order
operators. A reliability ladder degrades any failing program to the field-only special case, so the
full system stays at or above the field-only floor in aggregate. GeoSelect reaches 58.86~mIoU on
RRSIS-D and 55.27~mIoU on RISBench, more than twice and about $1.7$ times the prior
training-free results. Under a fixed detector and
segmenter, explicit execution improves RRSIS-D val by $+12.1$~mIoU over implicit
selectors. The discrete operators make the compositional constructions expressible that the field
algebra cannot and add a resolved gain on superlatives; the remaining headroom splits between
proposal recall and the program path, more of it on the program path for RISBench. The method targets single-instance expressions and
assumes an in-domain detector. Extending the field algebra to at-a-distance and metric-relative
readings, such as a tuned banded kernel for ``next to'' and scale-relative distance thresholds,
is left to future work. Executing an explicit spatial program brings training-free RRSIS
to about $93\%$ of the supervised specialist RMSIN on RRSIS-D while keeping every intermediate
decision inspectable.

%% file: tables/tab_mapping.tex
\begin{table}[!t]
\caption{Representative phrase-to-operator mappings from the fixed synthesis prompt (image polarity:
$X$ left-to-right, $Y$ top-to-bottom). The leaf \textsc{Select} for the head noun phrase is elided where
an operator applies to it, and the operands of \textsc{And}/\textsc{Or} share one \textsc{Select} query.
See Appendix~\ref{sec:appendix:schema} for the full EBNF.}
\label{tab:mapping}
\centering\footnotesize
\begin{tabular}{@{}>{\raggedright\arraybackslash}p{0.40\columnwidth}>{\raggedright\arraybackslash}p{0.55\columnwidth}@{}}
\toprule
Expression phrase & Operator skeleton (head \textsc{Select} elided) \\
\midrule
``a dam on the right'' & \textsc{Filter}(RIGHT, IMAGE) \\
``in the middle / central area'' & \textsc{Filter}(CENTER, IMAGE) \\
``in the lower left'' & \textsc{Filter}(LL, IMAGE) \\
``the rightmost airplane'' & \textsc{Argmax}(X) \\
``the leftmost airplane'' & \textsc{Argmin}(X) \\
``the bottommost tank'' & \textsc{Argmax}(Y) \\
``the topmost tank'' & \textsc{Argmin}(Y) \\
``the second ship from the left'' & \textsc{Nth}($n{=}1$, X) \\
``the ship to the left of the harbor'' & \textsc{Filter}(LEFT, \textsc{Select}(harbor)) \\
``a vehicle above the toll station'' & \textsc{Filter}(TOP, \textsc{Select}(toll station)) \\
``the rightmost of the three tanks on the left'' & \textsc{Argmax}(\textsc{RestrictCount}(\textsc{Filter}(LEFT, IMAGE), $k{=}3$), X) \\
``the building between the river and the road'' & \textsc{Relate}(BETWEEN, \textsc{Select}(river), \textsc{Select}(road)) \\
``the car near the building'' & \textsc{Relate}(NEAR, \textsc{Select}(building)) \\
``a light gray windmill'' & \textsc{Select}(``light gray windmill'') \\
``a stadium near the river and in the upper left'' & \textsc{And}(\textsc{Relate}(NEAR, \textsc{Select}(river)), \textsc{Filter}(UL)) \\
``a ship on the left or near the top'' & \textsc{Or}(\textsc{Filter}(LEFT), \textsc{Filter}(TOP)) \\
``a storage tank not near the building'' & \textsc{And}(\textsc{Select}(``storage tank''), \textsc{Not}(\textsc{Relate}(NEAR, \textsc{Select}(building)))) \\
\bottomrule
\end{tabular}
\end{table}